\documentclass[review, 3p, authoryear]{elsarticle}

\usepackage{amssymb,amsmath,amsthm}

\newtheorem{lemma}{Lemma}
\newtheorem{proposition}{Proposition}
\newtheorem{corollary}{Corollary}
\newtheorem{exmp}{Example}[section]

\usepackage{color}
\usepackage{tcolorbox}
\usepackage{hyperref}
\usepackage{booktabs}
\usepackage{multirow}
\usepackage{array}
\newcolumntype{L}[1]{>{\raggedright\let\newline\\\arraybackslash\hspace{0pt}}m{#1}}
\newcolumntype{C}[1]{>{\centering\let\newline\\\arraybackslash\hspace{0pt}}m{#1}}
\newcolumntype{R}[1]{>{\raggedleft\let\newline\\\arraybackslash\hspace{0pt}}m{#1}}
\usepackage{subcaption}
\usepackage{longtable}
\usepackage{arydshln}
\usepackage{soul}

\begin{document}


\newcommand\hlb[3][]{\todo[inline,caption={emptytext},
  size=\normalsize, backgroundcolor=yellow!70, bordercolor=yellow!70, noshadow, #1]{
    \begin{minipage}{
        \textwidth-4pt}#2
    \end{minipage}}
  \todo{\begin{spacing}{0.5}#3\end{spacing}}}

\newcommand{\hlc}[2]{\hl{#1}\todo{\begin{spacing}{0.5}#2\end{spacing}}}

\newcommand{\intodo}[2][]{\todo[inline, noshadow, caption={emptytext}, #1]{
    \begin{spacing}{1.0}\normalsize{#2}\end{spacing}}}

\newcommand{\sib}[1]{\textcolor{red}{\textbf{#1}}}
\newcommand{\tsib}[1]{\begin{tcolorbox}\textcolor{violet}{#1}\end{tcolorbox}}

\newcounter{sibcmntcounter}
\setcounter{sibcmntcounter}{1}
\long\def\symbolfootnote[#1]#2{\begingroup
  \def\thefootnote{\fnsymbol{footnote}}\footnote[#1]{#2}\endgroup}
\newcommand{\sibcmnt}[1]{{\small\textbf{
      \textcolor{violet}{(C.\arabic{sibcmntcounter})}}
    \let\thefootnote\relax\footnotetext{\textcolor{violet}
        {\scriptsize(C.\arabic{sibcmntcounter})~#1}}}
  \addtocounter{sibcmntcounter}{1}}

\newcommand{\red}[1]{\textcolor{red}{#1}}
\newcommand{\blue}[1]{\textcolor{blue}{#1}}
\newcommand{\magenta}[1]{\textcolor{magenta}{#1}}
\newcommand{\bm}[1]{\mbox{\boldmath{$#1$}}}
\newcommand{\rb}[1]{\raisebox{-1.5ex}[0cm][0cm]{#1}}
\newcommand{\HRule}{\noindent\rule{\linewidth}{0.5mm}}
\newcommand{\dsum}{\displaystyle\sum}
\newcommand{\veps}{\varepsilon}

\newcommand{\CA}{\mathcal{A}}
\newcommand{\CB}{\mathcal{B}}
\newcommand{\CC}{\mathcal{C}}
\newcommand{\CD}{\mathcal{D}}
\newcommand{\CG}{\mathcal{G}}
\newcommand{\CI}{\mathcal{I}}
\newcommand{\CJ}{\mathcal{J}}
\newcommand{\CK}{\mathcal{K}}
\newcommand{\CL}{\mathcal{L}}
\newcommand{\CN}{\mathcal{N}}
\newcommand{\CP}{\mathcal{P}}
\newcommand{\CS}{\mathcal{S}}
\newcommand{\CT}{\mathcal{T}}
\newcommand{\CX}{\mathcal{X}}
\newcommand{\ZZ}{\mathbb{Z}}
\newcommand{\RR}{\mathbb{R}}
\newcommand{\NN}{\mathbb{N}}
\newcommand{\II}{\mathbb{1}}

\newcommand{\va}{\bm{a}}
\newcommand{\vb}{\bm{b}}
\newcommand{\vc}{\bm{c}}
\newcommand{\vd}{\bm{d}}
\newcommand{\ve}{\bm{e}}
\newcommand{\vf}{\bm{f}}
\newcommand{\vg}{\bm{g}}
\newcommand{\vh}{\bm{h}}
\newcommand{\vp}{\bm{p}}
\newcommand{\vr}{\bm{r}}
\newcommand{\vt}{\bm{t}}
\newcommand{\vu}{\bm{u}}
\newcommand{\vv}{\bm{v}}
\newcommand{\vw}{\bm{w}}
\newcommand{\vx}{\bm{x}}
\newcommand{\vy}{\bm{y}}
\newcommand{\vz}{\bm{z}}
\newcommand{\zv}{\bm{0}}
\newcommand{\ov}{\bm{1}}

\newcommand{\vveps}{\bm{\veps}}
\newcommand{\veta}{\bm{\eta}}
\newcommand{\vxi}{\bm{\xi}}
\newcommand{\valpha}{\bm{\alpha}}
\newcommand{\vbeta}{\bm{\beta}}
\newcommand{\vgamma}{\bm{\gamma}}
\newcommand{\vtheta}{\bm{\theta}}
\newcommand{\vlambda}{\bm{\lambda}}
\newcommand{\vnu}{\bm{\nu}}
\newcommand{\vpi}{\bm{\pi}}
\newcommand{\vtau}{\bm{\tau}}

\newcommand{\mA}{\bm{A}}
\newcommand{\mB}{\bm{B}}
\newcommand{\mC}{\bm{C}}
\newcommand{\mD}{\bm{D}}
\newcommand{\mE}{\bm{E}}
\newcommand{\mF}{\bm{F}}
\newcommand{\mG}{\bm{G}}
\newcommand{\mH}{\bm{H}}
\newcommand{\mI}{\bm{I}}
\newcommand{\mL}{\bm{L}}
\newcommand{\mM}{\bm{M}}
\newcommand{\mP}{\bm{P}}
\newcommand{\mQ}{\bm{Q}}
\newcommand{\mR}{\bm{R}}
\newcommand{\mS}{\bm{S}}
\newcommand{\mU}{\bm{U}}
\newcommand{\mV}{\bm{V}}
\newcommand{\mX}{\bm{X}}

\newcommand{\tr}{^{\intercal}}
\newcommand{\ntr}{^{-\intercal}}
\newcommand{\inv}{^{-1}}

\newcommand{\gr}{\mbox{graph}}
\newcommand{\ra}{\rightarrow}
\newcommand{\la}{\leftarrow}
\newcommand{\Ra}{\Rightarrow}
\newcommand{\rra}{\rightrightarrows}
\newcommand{\ptr}{\marginpar{$\Leftarrow$}}

\newcommand{\pfxi}{\frac{\partial f(\vx)}{\partial x_i}}
\newcommand{\pfx}{\partial f(\vx)}
\newcommand{\pf}{\partial f}
\newcommand{\pxi}{\partial x_i}
\newcommand{\px}{\partial x}

\newcommand{\nfx}{\nabla f(\vx)}
\newcommand{\eps}{\epsilon}
\newcommand{\eg}{\textit{e.g.}}
\newcommand{\ie}{\textit{i.e.}}

\newcommand{\vsp}{\vspace{4mm}}
\newcommand{\vspp}{\vspace{8mm}}
\newcommand{\vsppp}{\vspace{12mm}}

\newcommand{\hsp}{\hspace{4mm}}
\newcommand{\hspp}{\hspace{8mm}}
\newcommand{\hsppp}{\hspace{12mm}}

\newcommand{\pr}[1]{\mathbb{P}\left(#1\right)}
\newcommand{\ex}[1]{\mathbb{E}\left[#1\right]}
\newcommand{\variance}[1]{\mbox{Var}\left(#1\right)}
\newcommand{\covar}[1]{\mbox{Cov}\left(#1\right)}
\newcommand{\C}[2]{\left(\begin{array}{c} #1 \\ #2 \end{array}\right)}

\newcommand{\maximize}{\mbox{maximize\hspace{4mm} }}
\newcommand{\minimize}{\mbox{minimize\hspace{4mm} }}
\newcommand{\subto}{\mbox{subject to\hspace{4mm}}}

\newenvironment{sibitemize}{
  \renewcommand{\labelitemi}{$\diamond$}
  \begin{itemize}
    \setlength{\parskip}{0mm}}
  {\end{itemize}}

\newcommand{\propnum}[2]{\vspace{3mm}
  \noindent {\sc Proposition #1}{\it #2} \vspace{3mm}}
\newcommand{\lemnum}[2]{\vspace{3mm}
  \noindent {\sc Lemma #1}{\it #2} \vspace{3mm}}
\newcommand{\thmnum}[2]{\vspace{3mm}
  \noindent {\sc Theorem #1}{\it #2} \vspace{3mm}}

\begin{frontmatter}

\title{Rule Generation for Classification: \\ Scalability, Interpretability, and Fairness}

\author[1]{Tabea E. R\"ober\corref{cor1}}\ead{t.e.rober@uva.nl}
\author[1]{Adia C. Lumadjeng}\ead{a.c.lumadjeng@uva.nl}
\author[2]{M. Hakan Aky\"uz}\ead{akyuz@ese.eur.nl}
\author[1]{Ş. İlker Birbil}\ead{s.i.birbil@uva.nl}

\affiliation[1]{organization={Amsterdam Business School, University of Amsterdam},
addressline={Plantage Muidergracht 12},
            postcode={1018TV}, 
            city={Amsterdam},
            country={The Netherlands}}

\affiliation[2]{organization = {Erasmus School of Economics, Erasmus University Rotterdam},
addressline = {Burgemeester Oudlaan 50},
                postcode = {3062 PA},
                city = {Rotterdam},
                country = {The Netherlands}}

\cortext[cor1]{Corresponding author}

\begin{abstract}
We introduce a new rule-based optimization method for classification with constraints. The proposed method leverages column generation for linear programming, and hence, is scalable to large datasets. The resulting pricing subproblem is shown to be NP-Hard. We recourse to a decision tree-based heuristic and solve a proxy pricing subproblem for acceleration. The method returns a set of rules along with their optimal weights indicating the importance of each rule for learning. We address interpretability and fairness by assigning cost coefficients to the rules and introducing additional constraints. In particular, we focus on local interpretability and generalize a separation criterion in fairness to multiple sensitive attributes and classes. We test the performance of the proposed methodology on a collection of datasets and present a case study to elaborate on its different aspects. The proposed rule-based learning method exhibits a good compromise between local interpretability and fairness on the one side, and accuracy on the other side.
\end{abstract}


\end{frontmatter}

\section{Introduction}\label{sec:introduction}

Classification is a supervised learning problem that assigns categorical labels, \textit{i.e.}, classes, to a target variable within a provided dataset. Decision trees (DTs) and decision rules are two related techniques commonly used in machine learning for classification tasks. A DT corresponds to a type of supervised learning algorithm that creates a tree-like model of decisions and their possible consequences. The tree is constructed by recursively partitioning the data into smaller subsets, based on the value of a feature, until the leaf nodes contain (almost) homogeneous subsets of the target variable. Decision rules, on the other hand, are sets of if-then statements that describe how the values of the input variables are used to determine the value of the target variable. These rules can be useful for gaining insights into the relationships among variables and for making predictions based on those relationships. One can consider a DT as a collection of dependent decision rules, since every leaf node of a DT corresponds to a decision rule.

In various application areas decision rules are considered to be interpretable by the decision makers. Such a rule consists of one or more conditions, and these conditions designate a class when all of them are satisfied. For example in binary classification, 
``\textbf{if} (\texttt{Last payment amount} is less than 100) \textbf{and} (\texttt{Months since issue date} is less than 9) \textbf{then} the credit risk is \textbf{good}''
is a rule that can be used to predict the credit risk of customers. When a sample satisfies this rule, then it receives a label corresponding to one of the two classes: ``\textbf{good}'' or ``\textbf{bad}''. In case a sample is covered by more than one rule, then the majority vote among the assigned labels may be used to determine the class of the sample.

In this work, we propose a rule-learning algorithm for classification with constraints based on mathematical programming. We present a general linear programming (LP) model for generating rules for multi-class classification. One advantage of using mathematical programming over off-the-shelf libraries containing popular algorithms (\textit{e.g.}, random forests, boosting methods, or neural networks) is its flexibility in adding specific constraints, such as those related to interpretability and fairness. Integrating such constraints in traditional machine learning (ML) techniques is not straightforward. 

\cite{bertsimas2017optimal} also emphasize this point and suggest that mathematical programming is highly flexible in solving ML problems. However, state-of-the-art approaches that address classification problems by solving mixed-integer linear programming (MILP) formulations quickly become intractable when there are thousands of integer variables \citep{bertsimas2017optimal,Aghaei-etal-2021StrongOptDTs,AlstonValidi2022MILOBinClass}. This significantly limits the practical use of MILPs for large datasets. Therefore, one of our motivations is to develop an LP-based approach that is both scalable enough to handle large datasets and flexible enough to incorporate constraints.

In our approach, each rule corresponds to a column in the mathematical programming model. When dealing with a large number of rules, our algorithm utilizes the column generation (CG) procedure to obtain the rules iteratively. This feature also connects our method to an earlier LP-based learning method, LPBoost by \cite{DemirizBennett2002LPBOOST}. LPBoost solves an LP that is similar to ours and learns using weak learners within a CG procedure. In that sense, our approach can be considered as a tailored LPBoost framework that is designed for decision rules considering interpretability and fairness aspects. Nevertheless, as we highlight in the following, our rule generation framework is different from the literature  and contributes to the existing literature in several ways. 

First, the LPBoost is characterized for binary classification problems. The works by \cite{GehlerNowozin2009FeatureCombiMulti} and \cite{Saffari-etal-2010MultiClassLPBoost} attempt to adapt the LPBoost into a multi-class setting, and both studies put in extra constraints. \cite{Saffari-etal-2010MultiClassLPBoost} build the LP formulation using a constraint for each sample-class pair similar to \textit{one-versus-rest} approach. In addition to extra terms and additional constraints, \cite{GehlerNowozin2009FeatureCombiMulti} also modify the constraints of the LPBoost. However, neither approach scales well, as adding an additional constraint per sample per class increases the computational burden of multi-class LPBoost. Our approach directly addresses multi-class classification without the need to define new constraints or resorting to \textit{one-versus-rest} approach on algorithmic level. 
Second, every weak learner of the LPBoost casts a vote for each sample in the constraints, whereas rules only apply to subset(s) of the data, namely when those samples satisfy the conditions in the corresponding rule. Our approach resolves this by associating each sample with a subset of rules for classification resulting in a sparser model since not all rules can cover a sample. In addition to a set of rules and their associated costs, our LP model also provides the optimal rule weights. These weights can be used to assign importance to each rule and are critical to locally interpret the resulting classification made for individual observations.
 Third, to the best of our knowledge, our work is the first to demonstrate that the resulting pricing subproblem to find a negative reduced cost rule is NP-hard when rule weights represent the rule lengths. To establish the computational complexity of the pricing subproblem, we require feature discretization and a one-vs-rest scheme to handle multiple classes. However, it is important to note that such feature discretization and one-vs-rest scheme are solely introduced for our theoretical analysis, and they are not restrictions imposed on the master problem formulation or on our rule generation framework by any means. To cope with the challenges of exactly solving the pricing subproblem, our rule generation framework leverages standard DTs with sample weights as a proxy pricing subproblem. Training DTs with sample weights is very fast and straightforward for most off-the-shelf ML libraries. The integration of such a fast heuristic for solving the pricing subproblem within our rule generation framework further enhances scalability.
Fourth, prior studies mentioned above primarily focus on accuracy performance and do not consider interpretability and fairness. Our method stands out by explicitly addressing these aspects which make it a more comprehensive approach for multi-class classification.  Furthermore, we put forth two metrics that can handle multiple classes and multiple protected groups for group fairness based on avoiding disparate mistreatment. These metrics are easily addressed as constraints in our general LP model.

The remainder of this work is organized as follows. Section \ref{sec:literature} discusses the relevant literature. Section \ref{sec:mat_mod} presents the methodology used and the rule generation algorithm proposed. In Section \ref{sec:num_exp}, we show our results on an extensive set of numerical experiments where the performance of our algorithm is tested against other benchmark studies. Conclusions are presented in Section \ref{section::conclusion}.

\section{Related Literature}\label{sec:literature}

The literature on mathematical optimization-based approaches (MOBAs) to solve classification problems is very rich. Therefore, we confine ourselves to the studies using DTs and rule-based methods that are most relevant to our proposed method. We refer to the works by \cite{Gambella2021OPTML}, \cite{Carrizosa-etal-2021MathematicalOpt} and \cite{Ignatiev-etal-2021ReasoningSurvey}, which provide excellent overviews of MOBAs for classification. 

\paragraph{Scalability} MILP formulations are prominent in rule-based learning on which our methodology also relies. \cite{MalioutovVarshney2013ExactRuleLearning} propose a rule-based binary classifier by solving an integer program that minimizes the number of rules for a boolean compressed sensing problem. \cite{WangRudin2015LearningOptimized} present a MILP formulation to collect decision rules for binary classification. 
\cite{DashGunlukWei2018BooleanCG} use a CG-based framework to find an optimal rule set for binary classification, where the objective is to balance the simplicity of classification rules and the algorithm's accuracy. For large instances, the pricing subproblem is either solved with time limits, or the columns are generated by a greedy heuristic. \cite{WeiDashGaoGunluk2019GeneralizedLinRuleModels} propose generalized linear rule models for binary classification using a similar CG framework as in \cite{DashGunlukWei2018BooleanCG}. \cite{CarrizosaIJOC10} also resort to a CG strategy for Binarized Support Vector Machines and show the scalibility of their approach while maintaining competitive classification accuracy.
\cite{MalioutovMeel2018MLIC-MaxSatRules} solve a MaxSAT formulation by constraint programming to construct interpretable classification rules. \cite{GhoshMeel2019_IMLI_IncrementalMaxSATrules} also propose a framework based on MaxSAT formulation that can be applied to binary classification problems with binary features to minimize the number of generated rules and the number of misclassified samples.

Decision rules (or rule sets) show similarity with DTs in the sense that every leaf node of a DT can be translated into a single rule by following the decision path starting from the root node. 
\cite{bertsimas2017optimal} aim to find an optimal DT for classification by formulating a MILP problem where single and multi-dimensional half-spaces recursively divide the feature space into disjoint areas during its construction. \cite{VerwerZhang2019Learning} propose a MILP formulation based on binary encoding of features to learn optimal DTs with a predefined depth. \cite{McTavish-etal-2021} introduce a method for fast sparse DT optimization via smart guessing strategies applicable to any optimal branch-and-bound-based DT algorithm.
Other notable works considering DTs include \cite{FiratCrognierGabor-etal-2020CG,Aghaei-etal-2021StrongOptDTs,Gunluk-etal-2021OptDTs,AlstonValidi2022MILOBinClass}, and \cite{Demirovic-etal-2022MurTree}. These studies depend on solving MILPs or enumerative search strategies, and thus, suffer from intractability on large datasets due to their long-running time requirements. More recently, \cite{PATEL2024106579} offer a column generation scheme that improves the study by \cite{FiratCrognierGabor-etal-2020CG} to approximate optimal DTs and mitigate excessive solution time requirements. \cite{BLANQUERO2020255} resort to continuous optimization to mitigate exploding computation times, and aim to find sparse optimal classification trees using oblique cuts. Such oblique decision trees may encounter interpretability challenges since there can be more than one feature for each condition in its rules. For decision rules that we use, this is strictly equal to one feature per condition in a rule. Unlike the aforementioned studies, our methodology is based on an LP formulation to achieve scalability. 

\paragraph{Interpretability}
To address interpretability, the MOBAs above mostly rely on the comprehensibility of shallow DTs or simplicity of rules to human understanding. The latter is often considered easier to understand than a DT, since even a moderate-depth DT can be relatively complex \citep{Furnkranz1999RuleLearning}. Many studies directly address interpretability, for example aiming to induce sparsity and limit the number of rules \citep[\textit{e.g.,}][]{LakkarajuBachLeskovec2016InterpretableDecisionSets, WangRudinVelezLiu2017BayesianRuleSet,WangWangGeng-et-al2019StackingBasedEnsemble,Guidotti-etal-2018LORE, BertsimasJaillet2019_Price_Interpretability,BLANQUERO2020255,ProencaLeeuwen2020ClassyII,Lawless-etal-2021FairBoolean,YangLeeuwen2022ProbabilisticRuleSets}. This perspective is also called \textit{global} interpretability by \cite{Molnar2022_IntML}. Although global interpretability can provide insight into the characteristics of the data in general, it is of limited practical usage when individualized recipes are to be provided at the sample level. The latter is termed \textit{local} interpretability \citep{Molnar2022_IntML} and relevant to recommend patient-specific medical recipes or loan application approval of consumers that require individualized explanations. Unlike the existing MOBAs mentioned, we also concentrate on local interpretability to fill the gap in the literature. 

\paragraph{Fairness}
Fairness is often formulated using additional constraints to avoid discrimination over disadvantaged groups in the data and receives growing attention in the MOBAs literature. \citet{Lawless-etal-2021FairBoolean} extend the rule learning framework of \citet{DashGunlukWei2018BooleanCG} to take into account fairness for binary classification. Likewise, \cite{JoAghaei-etal-2022OptFairDTs} incorporate different group fairness notions by modifying the MILP formulation from \cite{Aghaei-etal-2021StrongOptDTs} for binary classification. We generalize fairness notions beyond binary classification to encompass multiple classes and also extend them to involve multiple sensitive attributes.

\section{Mathematical Programming Models}
\label{sec:mat_mod}

Consider a training dataset $\CD = \{(\vx_i, y_i) : i \in \CI\}$ for a classification problem, where $\CI$ is the index set of samples, $\vx_i \in \RR^d$ and $y_i$ denote the vector of features and the label of sample $i \in \CI$, respectively. We define a rule as a conjunction of conditions that partitions the feature space of the samples. These conditions are often called as \textit{literals} in Boolean formulae literature for concept learning, and the rule itself corresponds to a \textit{term} in a Disjunctive Normal Form (DNF) \citep[see][]{Rivest1987Boolean,Kearns-etal-1994Boolean}.

The main idea of our proposed learning approach is to obtain a set of rules indexed by $\mathcal{J}$ and the corresponding nonnegative rule weights $w_j$, $j \in \mathcal{J}$ so that the prediction for each sample is given as a \textit{weighted} combination of the rule predictions. A sample $i \in \CI$ is covered by a rule $j \in \mathcal{J}$ if and only if the sample satisfies all conditions in the rule. To this end, we construct LP models that minimize the total loss and the overall cost of using the rules. As we shall see next, working with LP models gives us a valuable advantage for scaling our approach to much larger datasets and considering multi-class problems. Moreover, we can easily extend our models with additional constraints in our training process without any need for pre- or post-processing of input or output data.

\begin{figure}
    \centering
    \includegraphics[width=0.8\textwidth]{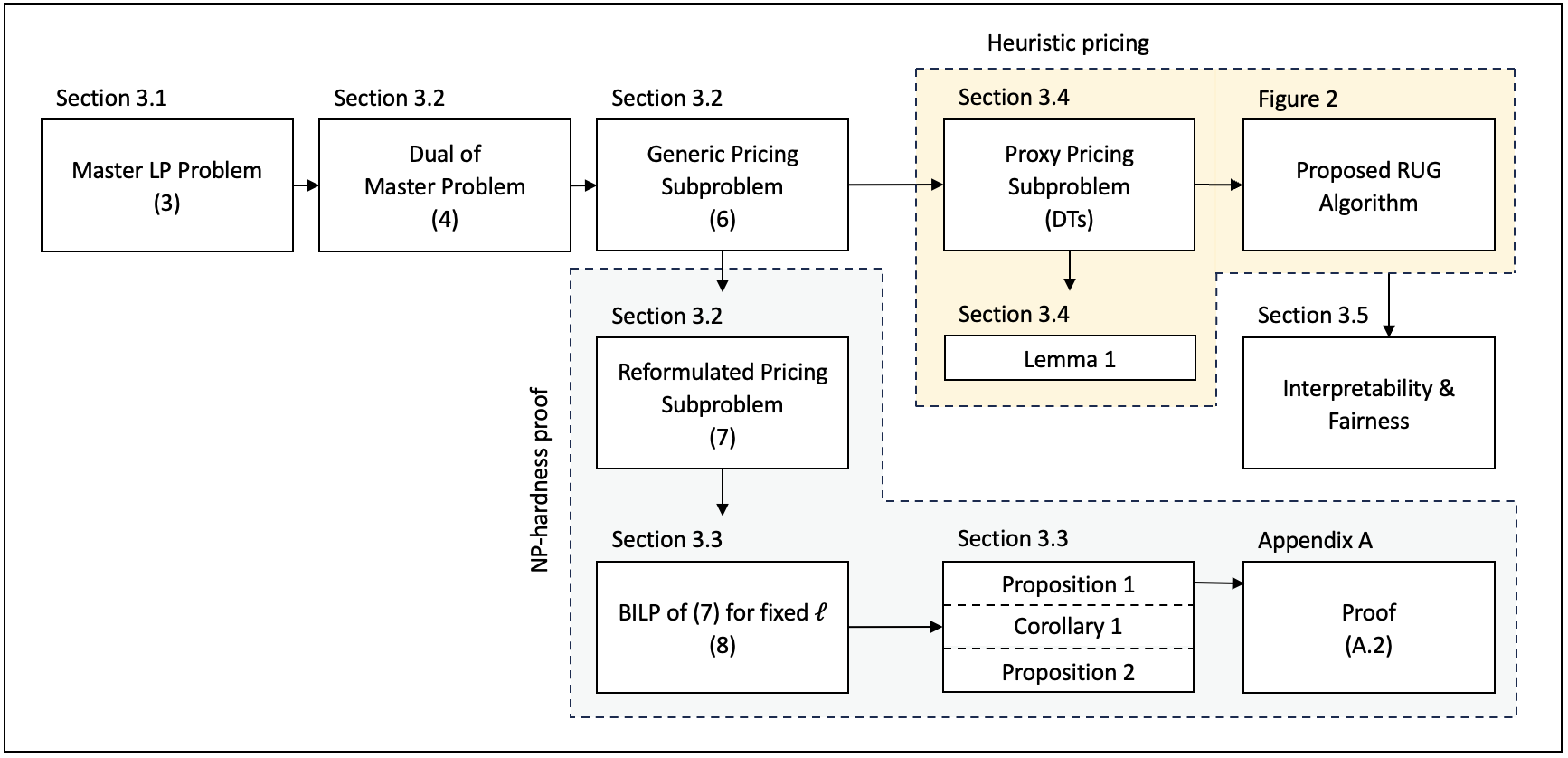}
    \caption{Overview of models discussed in Section \ref{sec:mat_mod}.}
    \label{fig:models}
\end{figure}

The subsequent segment of this section consists of five parts. We first introduce in Subsection \ref{sec:master_mod} the master problem that is used for the classification problem. Subsection \ref{section:rulegen} elucidates the rule generation scheme used for the column generation (CG) procedure. The resulting pricing subproblem structure of the CG, its complexity, and our proxy pricing subproblem approach are presented in subsections \ref{sec:PSP} and \ref{sec:proxyPSP}. Finally, we devote Subsection \ref{section::sub::interpretability} to addressing the interpretability and incorporating fairness constraints into our model. In Figure \ref{fig:models} we visualize how the models we discuss in this section relate to each other to give a concise overview.

\subsection{Master Problem}
\label{sec:master_mod}

Suppose that the problem consists of $K$ classes. In multi-class classification, we need a way to represent a sample belonging to one of $K$ different classes. To do so, we define a vector-valued
mapping $\mathbf{y}_i \in \mathbb{R}^K$ as in \cite{Multi_Class_Adaboost2009}, which enables us to apply multi-class loss functions and to formulate the optimization problem in a unified manner.
That is, if $y_i=k$, then
\begin{equation}
  \label{eqn:vecmap}
  \vy_i = (-\tfrac{1}{K-1}, -\tfrac{1}{K-1}, \dots, 1, \dots, -\tfrac{1}{K-1})^{\intercal},
\end{equation}
where the value one appears only at the $k^{th}$ component of the vector. A given rule $j \in \mathcal{J}$ assigns the vector $\mR_j(\vx_i) \in \RR^K$ to input $\vx_i$, only if the rule covers sample $i \in \CI$. This vector is also formed in the same manner as in \eqref{eqn:vecmap} and represents the rule $j$'s prediction for sample $i$. The final prediction of sample $i$ is then made using a weighted average of all rule predictions, $\mR_j(\vx_i)$'s, and their associated weights, $w_j$'s. Now, the prediction vector for sample $i \in \CI$ becomes
\[
 \hat{\vy}_i(\vw) = \sum_{j \in \mathcal{J}} a_{ij}\mR_j(\vx_i)w_j,
\]
where $\vw = (w_j: j \in \mathcal{J})\tr$ represents the vector of rule weights, and $a_{ij} \in \{0, 1\}$ indicates whether rule $j \in \mathcal{J}$ covers sample $i \in \CI$ or not; \textit{i.e.}, $a_{ij} = 1$ or $a_{ij} = 0$. The predicted class of sample $i$ is determined by assigning it to the highest element (to the corresponding class) in the prediction vector $\hat{\vy}_i(\vw)$. We break ties arbitrarily, namely, we assign the class of a sample randomly among the highest value elements in its prediction vector.

\begin{exmp}\label{example1}
Consider a classification problem with $K = 3$ classes: $1$, $2$, and $3$. Each sample $\vx_i$ has a class label $\vy_i \in \RR^K$, where
\[
\vy_i =
\begin{cases}
(1, -\tfrac{1}{2}, -\tfrac{1}{2})^{\intercal}, & \text{if } y_i = 1, \\
(-\tfrac{1}{2}, 1, -\tfrac{1}{2})^{\intercal}, & \text{if } y_i = 2, \\
(-\tfrac{1}{2}, -\tfrac{1}{2}, 1)^{\intercal}, & \text{if } y_i = 3.
\end{cases}
\]
Suppose we have a set of four rules such as those generated by our proposed algorithm. The four rules, $R_1$, $R_2$, $R_3$, $R_4$, assign prediction vectors $\mR_j(\vx_i) \in \RR^K$, as follows:
\[
\mR_1(\vx_i) = \begin{bmatrix} 1 \\ -\frac{1}{2} \\ -\frac{1}{2} \end{bmatrix}, \quad
\mR_2(\vx_i) = \begin{bmatrix} -\frac{1}{2} \\ 1 \\ -\frac{1}{2} \end{bmatrix}, \quad
\mR_3(\vx_i) = \begin{bmatrix} 1 \\ -\frac{1}{2} \\ -\frac{1}{2} \end{bmatrix}, \quad
\mR_4(\vx_i) = \begin{bmatrix} -\frac{1}{2} \\ -\frac{1}{2} \\ 1 \end{bmatrix}.
\]
The rule weights are $\vw = [0.6, 0.4, 0.5, 0.3]^\intercal$. Assume sample $\vx_1$ satisfies $R_1$, $R_3$, and $R_4$, but not $R_2$, so its coverage vector is $\bm{a}_1 = [1, 0, 1, 1]^\intercal$.
The prediction vector is then given by
\[
\hat{\vy}_1(\vw) = \sum_{j \in \mathcal{J}} a_{1j}\mR_j(\vx_1)w_j = \begin{bmatrix} 1 \\ -\frac{1}{2} \\ -\frac{1}{2} \end{bmatrix} 0.6 + \begin{bmatrix} 1 \\ -\frac{1}{2} \\ -\frac{1}{2} \end{bmatrix} 0.5 + \begin{bmatrix} -\frac{1}{2} \\ -\frac{1}{2} \\ 1 \end{bmatrix} 0.3 = \begin{bmatrix} 0.95 \\ -0.7 \\ -0.25 \end{bmatrix}.
\]
Finally, the predicted class corresponds to the maximum component of $\hat{\vy}_1(\vw)$. That is 
\[
\arg \max_k \{\hat{\vy}_1(\vw)\} = 1.
\]
\end{exmp}

In order to evaluate the total classification error, we use the \textit{hinge loss} function and define
\begin{equation}
    \CL(\hat{\vy}_i(\vw), \vy_i) = \max\{1 - \kappa\hat{\vy}_i(\vw)\tr\vy_i, 0\}, \label{eq::loss-function}
\end{equation}
where $\kappa = (K-1)/K$. Next, we introduce the auxiliary variables $v_i$, $i \in \CI$ standing for $v_i \geq \CL(\hat{\vy}_i(\vw), \vy_i)$. This condition ensures that $v_i =0$ only when the term $\kappa\hat{\vy}_i(\vw)\tr\vy_i \geq 1$ in \eqref{eq::loss-function}, which corresponds to correct classification for sample $i \in \CI$.
However, a positive value of $v_i$ does not necessarily indicate a misclassification but rather it shows the extent to which the prediction deviates from the true label, based on the weighted combinations of rule predictions. When $v_i$ increases, it shows increasing difficulty of correctly classifying a sample. We illustrate how $v_i$ values are calculated using \eqref{eq::loss-function} in the following example which continues from Example \ref{example1}.

\begin{exmp}\label{example2}
Consider the label vector $\vy_1$ of sample $\vx_1$ for three cases i) $y_1 = 1$, ii) $y_1 = 2$ and iii) $y_1 = 3$. We perform the operations stated in \eqref{eq::loss-function} as follows.

i) Case $y_1 = 1$: $\vy_1 = (1, -\tfrac{1}{2}, -\tfrac{1}{2})^{\intercal}$ then $\kappa \hat{\vy}_1(\vw)\tr\vy_1 = \tfrac{2}{3}(0.95, -0.7, -0.25) (1, -\tfrac{1}{2}, -\tfrac{1}{2})^{\intercal} = 0.95$ This implies that $v_1 = 1-0.95 = 0.05$ by \eqref{eq::loss-function} if the sample belongs to the first class.

ii) Case $y_1 = 2$: $\vy_1 = (-\tfrac{1}{2}, 1, -\tfrac{1}{2})^{\intercal}$ then $\kappa \hat{\vy}_1(\vw)\tr\vy_1 = \tfrac{2}{3}(0.95, -0.7, -0.25) (-\tfrac{1}{2}, 1, -\tfrac{1}{2})^{\intercal} = -0.7$ This implies that $v_1 = 1-(-0.7) = 1.7$ by \eqref{eq::loss-function} if the sample belongs to the second class.

iii) Case $y_1 = 3$: $\vy_1 = (-\tfrac{1}{2}, -\tfrac{1}{2}, 1)^{\intercal}$ then $\kappa \hat{\vy}_1(\vw)\tr\vy_1 = \tfrac{2}{3}(0.95, -0.7, -0.25) (-\tfrac{1}{2}, -\tfrac{1}{2}, 1)^{\intercal} = -0.25$ This implies that $v_1 = 1-(-0.25) = 1.25$ by \eqref{eq::loss-function} if the sample belongs to the third class.
\end{exmp}
As exemplified in Example \ref{example2}, $v_i \geq 1$ implies that the prediction vector $\hat{\vy}_i(\vw)$ and the label vector $\vy_i$ of sample $i$ are not in agreement at all, and thus, sample $i$ is misclassified. On the other hand, when $0< v_i < 1 $ the $v_i$ value shows how certain a rule from its decision. It can be expected when $v_i$ value is closer to 1 for sample $i$, it might be misclassified by the rule. Likewise, a $v_i$ value that is close to zero is likely to indicate correct classification yet only $v_i=0$ implies a correct classification. Such a property is also common in support vector machines \citep[see][]{Hastie2009ElementsSL}. 

Now, we are ready to present the LP model of our master problem by using our auxiliary variables $v_i$'s:
\begin{equation}
  \label{eq:master_model}
    \begin{array}{lll}
        \minimize & \lambda \sum_{j \in \mathcal{J}} c_j w_j + \sum_{i \in \CI} v_i & \\[2mm]
        \subto & \sum_{j \in \mathcal{J}} \hat{a}_{ij}w_j + v_i \geq 1, & i \in \CI, \\[2mm]
        & v_i \geq 0, & i \in \CI, \\[2mm]
        & w_j \geq 0, & j\in\mathcal{J},
    \end{array}
\end{equation}
where $\hat{a}_{ij} = \kappa a_{ij}\mR_{j}(\vx_i)\tr \vy_i$ is a measure of classification accuracy of rule $j$ for sample $i$ given that the sample is covered by the rule, and the coefficients $c_j \geq 0$, $j \in \mathcal{J}$ stand for the costs of rules. 
For example, these cost coefficients can reflect the length of the rule (\textit{i.e.}, the number of conditions used) for sparsity or they may be simply set to one. The objective function shows the trade-off between the accuracy and the cost of using rules. Since these two summation terms are in different units, the hyperparameter $\lambda \geq 0$ is used for scaling. 

\subsection{Rule Generation}
\label{section:rulegen}

The LP model \eqref{eq:master_model} of our master problem operates with a given set of rules, $\mathcal{J}$. Notice that $\mathcal{J}$ contains exponentially many rules due to all possible splitting combinations (namely \textit{partitions}) of the features. The number of such partitions of the feature space based on the training dataset $\CD$ can be counted using \textit{Bell numbers} \citep[see][]{Liu2022bsnsing}, which grow quickly with the number of features and their associated levels. This aspect shows the combinatorial structure of the problem and prohibits us to generate the set of rules explicitly as its size becomes extremely large. Hence, we resort to an alternative way to generate the desired rules iteratively as we explain next.

In the master problem, the rules correspond to columns, and obtaining them in an iterative scheme leads to the well-known column generation (CG) procedure in optimization \citep{DesaulniersDesrosiersSolomon2006ColumnGeneration}. At each iteration of this approach, an LP model is constructed with a subset of the columns (\textit{column pool}). 
This model is called the \textit{restricted master problem}. After solving the restricted master problem, a dual optimal solution is obtained. Then, using this dual solution, a \textit{pricing subproblem} (PSP) is solved to identify the columns with negative \textit{reduced costs}. These columns are the only candidates for improving the objective function value when they are added to the column pool. The next iteration continues after extending the column pool with the negative reduced cost columns.

In order to apply CG to our problem, we define a subset of rules $\mathcal{J}_t \subset \mathcal{J}$ at iteration $t$ and form the restricted master problem by replacing $\mathcal{J}$ with $\mathcal{J}_t$ in our LP models. Let us denote the dual variables associated with the first set of constraints in \eqref{eq:master_model} by $\beta_i$, $i \in \CI$. In vector notation, we use boldface font as $\vbeta$. Then, the \textit{dual restricted master problem} at iteration $t$ becomes
\begin{equation}
  \label{RMP_d_class}
  \begin{array}{lll}
    \maximize & \sum_{i \in \CI} \beta_i & \\[2mm]
    \subto    & \sum_{i \in \CI} \hat{a}_{ij}\beta_i  \leq \lambda c_j, & j \in \mathcal{J}_t, \\[2mm]
              &  0 \leq \beta_i \leq 1, & i \in \CI.
  \end{array}
\end{equation}
If we denote an optimal dual solution at iteration $t$ by $\bm{\beta}^{(t)}$, then improving the objective function value of problem \eqref{RMP_d_class} requires finding at least one rule $j' \in \mathcal{J} \setminus\mathcal{J}_t$ such that
\begin{equation}
  \label{redcostcheck}
  \bar{c}_{j'} = \lambda c_{j'} - \sum_{i \in \CI} \hat{a}_{ij'}\beta_i^{(t)} < 0,
\end{equation}
where $\bar{c}_{j'}$ is also known as the \textit{reduced cost} of column $j'$. In fact, this
condition simply checks whether $j' \in \mathcal{J} \setminus \mathcal{J}_t$ violates the dual feasibility. To find those rules with negative reduced costs, we formulate the pricing subproblem for classification as
\begin{equation}
  \label{PSP_class}
  \underset{j \in \mathcal{J} \setminus \mathcal{J}_t}{\min}\left\{\lambda c_j - \sum_{i \in \CI} \hat{a}_{ij}\beta_i^{(t) }  
  \right\}.
\end{equation}
Note that the second term measures the \textit{weighted} classification error for all those points classified by rule $j \in \mathcal{J} \setminus \mathcal{J}_t$. The weight for each sample $i \in \CI$ is given by the corresponding dual optimal solution, $\beta_i^{(t)}$. We will use this observation to devise a method to solve our PSP.

When the PSP does not return any rule with a negative reduced cost, then we have an optimal solution to the master problem with the current set of rules, $\mathcal{J}_t$. Otherwise, we have at least one rule with a negative reduced cost. After adding one or more rules with negative reduced costs to $\mathcal{J}_t$, we proceed with $\mathcal{J}_{t+1}$ and solve the restricted master problem or, equivalently, its dual \eqref{RMP_d_class}. It is important to note that the sole purpose of solving the PSP is to return a subset of rules with negative reduced costs. 

To further discuss the complexity of the PSP in \eqref{PSP_class}, we assume in the subsequent analysis part that $c_j$ is integer for all $j \in \CJ$ and varies from one to $L$. A natural example is the number of features (also related with the rule length) which is bounded by the dimension of the input space. This allows us to partition the rule set $\mathcal{J} \setminus \mathcal{J}_t$ into disjoint subsets. Let $\hat{\mathcal{J}}^{(\ell)}_{t}$ be the partition $\ell=1, \ldots, L$ that is defined as $\hat{\mathcal{J}}^{(\ell)}_{t} = \{j \in \mathcal{J} \setminus \mathcal{J}_t: c_j = \ell \}$. Here $c_i = \ell$ implies a subset of rules having their associated cost equal to $\ell$. We should note that this structure is similar to a two-phase mathematical programming formulation \citep[see][]{Bertsimas2021Unified} as the second term $\sum_{i \in \CI} \hat{a}_{ij}\beta_i^{(t)}$ in \eqref{PSP_class} is dependent on the rule cost value $c_j$. By construction, $\hat{\mathcal{J}}^{(\ell)}_{t} \cap \hat{\mathcal{J}}^{(\bar{\ell})}_{t} = \varnothing $ for $\ell \neq \bar{\ell} = 1, \ldots, L$. Using this partitioning structure, we obtain 
\begin{equation}
  \label{PSP_class_partitioned}
  \underset{\ell = 1, \ldots, L}{\min} \left\{  \underset{j \in \hat{\mathcal{J}}^{(\ell)}_{t}}{\min}\left\{\lambda c_j - \sum_{i \in \CI} \hat{a}_{ij}\beta_i^{(t) }  
  \right\} \right\}.
\end{equation}
Observe that $\mathcal{J} \setminus \mathcal{J}_t = \hat{\mathcal{J}}^{(1)}_{t} \bigcup \hat{\mathcal{J}}^{(2)}_{t} \bigcup \ldots \bigcup \hat{\mathcal{J}}^{(L)}_{t} $. Thus, \eqref{PSP_class} and \eqref{PSP_class_partitioned} are equivalent to each other. This fact is used later for our analysis. 

In what follows, we discuss details of the PSP in \eqref{PSP_class_partitioned} and present an analysis of its complexity to demonstrate the difficulty of exactly solving it. Note that although the PSP in \eqref{PSP_class} is generic, it is not easy to formulate a tractable mathematical programming formulation (\textit{i.e.}, linear or integer formulations without non-linearities) to our knowledge. 
To that end, we first reformulate the PSP in \eqref{PSP_class_partitioned} as a binary integer linear programming (BILP) formulation. We impose additional assumption that all features are discretized. Once the complexity analysis is presented, we then continue with a proxy PSP, which is used as heuristic rule generation in our CG procedure to bypass the computational burden of solving the proposed PSP formulation exactly. For the sake of completeness, we also compare these two approaches, \textit{i.e.}, solving the proxy PSP and the PSP formulation in our experiments (see Section \ref{sub:proxy_exact}). However, the BILP formulation in the following part is solely used for theoretical analysis and does not have an impact on the generality of our master problem \eqref{eq:master_model} with respect to its multi-class and/or encoding-free structure.

\subsection{Pricing Subproblem}
\label{sec:PSP}

We formulate the PSP stated in \eqref{PSP_class_partitioned} as a binary integer linear program (BILP). To that end, we focus on the case where we have a target class $k \in \CK$ to predict, and it is possible to consider each class in a \textit{one-versus-rest} setting without loss of generality. At first glance, it might seem to be a simplification to formulate the PSP as a binary classification problem using \textit{one-versus-rest} instead of directly modelling it as a multi-class problem. There are two primary reasons for this choice. First, the multi-class problem requires introducing extra integer variables to represent the classes associated with a rule. Second, the multi-class problem leads to a nonlinear integer formulation. As shown in the subsequent, we take advantage of two classes (\textit{one-versus-rest}) to fix a rule's prediction as positive for the target (\textit{i.e}., positive) class, and thus, represent the PSP as a BILP. Observe that \textit{one-versus-rest} setting can be achieved in polynomial time considering each class $k \in \CK$. Similarly, we devise a one-hot encoding scheme as described in the works by \cite{Hastie2009ElementsSL} and \cite{DashGunlukWei2018BooleanCG}. The features are transformed into a binary form. For categorical features, a binary variable per category and their negations are created indicating whether a sample belongs to a category or not. For numerical features, threshold values corresponding to the deciles of the dataset $\CD$ are used to define binary variables similarly \citep[for further details see][]{DashGunlukWei2018BooleanCG}. The dataset $\CD$ is also represented using such a binarization of features that can be achieved in polynomial time of the dataset size. We should emphasize that our LP framework is generic and the master problem \eqref{eq:master_model} does not necessitate the discretization of features. Here, our aim is to provide a comprehensive analysis of the PSP and to show theoretical properties associated with its structure.

Before delving into the details of the PSP formulation, we first introduce some additional notation. For simplicity, we remove the iteration index $t$ and the rule index $j$ from the PSP formulation as we are seeking an optimal rule $j^*$. Let $a_i$ and $z_p$ be binary variables, where $a_i$ takes the value of one if sample $i \in \CI$ is covered by the rule, and zero otherwise. Similarly, $z_p$ takes the value of one if feature $p \in \CP$ is not included in the rule, and zero otherwise. Here, $\CP$ stands for the set of features after binarization. Notice that, the variable $z_p$ is more relevant for our purposes. When all features are included, this implies that a rule spans the entire feature space. Features need to be excluded accordingly to obtain conditions in an if statement of the rule. We further define the binary variable $b_i$, by substituting $b_i = 1-a_i$, which is equal to one when the sample is not covered by the rule, zero otherwise. Although, we use $b_i$ in our BILP formulation to be concise, we still need $a_i$ for clarity in the following discussions. The parameter $x_{ip}$ has a value of one when sample $i$ satisfies feature $p$ and zero otherwise. Let $\CP_i$ denote the set of features needed in the rule to cover sample $i$, \textit{i.e.}, $\CP_i = \{ p\in \CP: x_{ip}=1 \}$. The samples $i \in \CI$ are divided into two subsets as $\CI^{+}$ and $\CI^{-}$ where the former set of samples has the same label with the target class $k$ and the latter set of samples is from other classes, \textit{i.e.}, $\CK \setminus \{ k \} $. Now, we can represent the label of sample $i \in \CI^{+}$ (or $i \in \CI^{-}$), \textit{i.e.}, when $y_i=k$ (or $y_i \neq k$), with $\vy_i = (1, -1)^{\intercal}$ (or $\vy_i = (-1, 1)^{\intercal}$). Similarly, prediction $\mR(\vx_i)$ of the rule for sample $i$ becomes $\mR(\vx_i) = (1, -1)^{\intercal}$ and $\mR(\vx_i) = (-1, 1)^{\intercal}$ when sample $i$ is covered, $a_i =1$, and not covered, $a_i =0$, respectively. Thus, omitting the rule index $j^*$, $\hat{a}_{i}$ in \eqref{PSP_class_partitioned} for sample $i$ reduces to $\hat{a}_{i} = \kappa a_{i}\mR(\vx_i)\tr \vy_i = a_i$ for $i \in \CI^{+}$ and $\hat{a}_{i} = -a_i$ for $i \in \CI^{-}$. After rearrangements using \eqref{PSP_class_partitioned}, the resulting problem for $\ell$ is presented as follows:
\begin{equation}
  \label{PSP_explicit2}
  \begin{array}{llll}
  d(\ell) = & \maximize & \sum_{i \in \CI^-} b_i \beta_i -\sum_{i \in \CI^+} b_i \beta_i + C \\[2mm]
    & \subto & \sum_{p \in \CP_i} z_p \geq b_i & i \in \CI, \\[2mm]
    & & z_p \leq b_i & i \in \CI, p \in \CP_i, \\[2mm]
    & & \sum_{p \in \CP} z_p = \ell, & \\[2mm]
    & & b_i \in \{ 0, 1 \}, & i \in \CI, \\[2mm]
    & & z_p \in \{ 0, 1 \}, & p \in \CP.
  \end{array}
\end{equation}
Here, we are interested to find the rule corresponding to $\underset{\ell = 1, \ldots, L}{\max} \left\{ d(\ell) \right\}$. The constant term represented with $C$ is as follows $C = \sum_{i \in \CI^+} \beta_i - \sum_{i \in \CI^-} \beta_i - \lambda \ell$ and should be added to the objective in \eqref{PSP_explicit2}. Clearly, such a constant does not change the optimal solution. 
The objective function simultaneously minimizes the weighted number of samples from target class $k$ and maximizes the weighted number of samples from other classes not covered by the rule. Notice that, $\ell$ is fixed in the third term, and $\sum_{p \in \CP} z_p = \ell$ in the third constraint is the associated cost of rule that counts the number of features excluded, which defines the number of conditions (\textit{i.e.}, rule length or number of literals) used for classification. This constraint often arises as a requirement on the length (or the number of literals ) of a rule to address interpretability where lower values for $\ell$ are preferred. In that respect, $\ell$ is treated as a hyperparameter in our setting. Also, observe that, when binarization of the data is performed, each category (or a split value for numerical features without loss of generality) is associated with two binary variables: one to define a category of a feature and the other for its negation \citep[see][]{DashGunlukWei2018BooleanCG}. Then, a condition is only defined when either the binary variable for the category or for its negation is not selected by the model (\textit{i.e.}, $z_p=1$). Otherwise, the rule either satisfies that category for all samples or covers no samples, and thus, there is no condition at all. Therefore, the number of conditions in a rule is calculated by simply counting the binary variables that are not selected in the formulation. 
The first constraint set in \eqref{PSP_explicit2} enforces the sample to be covered by the rule (\textit{i.e.}, $b_i=0$) when the term on the left-hand side is zero for all features (\textit{i.e.}, for all features $p \in \CP_i$ needed to cover the sample with $x_{ip} = 1$) and thus $z_p$'s are zero for all features $p \in \CP_i$ that are not to be excluded from the rule. Otherwise, the first constraint set is redundant. This implies that sample $i$ is covered, \textit{i.e.}, $a_i = 1$, if and only if there is no case a necessary feature $p$ is not selected for $i$. 
The second constraint set complements this and enforces $b_i = 1$ when $z_p = 1$ with $x_{ip}=1$ for some feature $p \in \CP_i$ which is sufficient to conclude that the sample is not covered by the rule. The last two constraint sets impose binary restrictions on the variables.  

\begin{proposition}\label{lem:psp}
    The PSP in \eqref{PSP_explicit2} is NP-hard.
\end{proposition}

The proof of Proposition \ref{lem:psp} is given in \ref{app::hardness_proof}. Clearly, there can be at most $|\CP|$ features (not) selected in a rule and the optimum can be found in polynomial time of the number of features by solving for $\ell = 1, \ldots, L \equiv |\CP| $, and choosing the minimum reduced cost rule among them. However, number of not selected features $\ell$ becomes an explicit constraint for interpretability purposes and one aims to keep rules simple. So, in practice, $\ell$ is imposed as a hyperparameter and a few values of it need to be cross-validated depending on the application domain or experts' preferences. 
We should also note that the PSP in \eqref{PSP_explicit2} has a similar structure with the maximum monomial aggreement problem (MMAP) as described in \cite{eckstein2012improved}, who have also shown that MMAP is NP-hard. A similar complexity result is also supported by the analysis of \cite{ben2003difficulty}. Nevertheless, the PSP in \eqref{PSP_explicit2} is slightly different than the MMAP in defining the coverage of a sample by a rule. The set of constraints is not the same with the MMAP \citep[for details, see the work by][]{eckstein2012improved}. For completeness, we have opted for giving our own proof which is customized to the current setting. The following corollary is a direct implication of Proposition \ref{lem:psp}. Its proof directly follows when we set $\beta_i = 1$ for each sample $i \in \CI$.

\begin{corollary}\label{lem:psp:corollary}
    Given $\ell$, constructing an optimal classification rule is NP-hard.
\end{corollary}

It is also possible to derive optimality bounds on \eqref{eq:master_model} based on the reduced cost expression in \eqref{PSP_class_partitioned}. Given $\ell$, let ${Z^{(\ell)}}^*$, $\overline{Z^{(\ell)}}$, and ${Z^{(\ell)}_k}^*$ denote, respectively, the optimum objective value of the master problem \eqref{eq:master_model}, the best objective value found so far for the master problem \eqref{eq:master_model}, and the optimum value for the PSP \eqref{PSP_explicit2} solved to find a rule for class $k$ for the corresponding $\ell$. 
Note that, \eqref{PSP_explicit2} is solved $K$ times once for each class $k = 1, \ldots, K$ to generate rules to add in the master problem. Then, the following proposition provides a lower bound $Z_{LB}$ on ${Z^{(\ell)}}^*$. 

\begin{proposition}\label{lem:optbound}
    $Z_{LB} = \overline{Z^{(\ell)}} - \underset{k = \{ 1, \ldots, K \} }{\max} \{ {Z^{(\ell)}_k}^* \} $ is a lower bound on ${Z^{(\ell)}}^*   $.
\end{proposition}

The proof is omitted and follows from LP duality. Each ${Z^{(\ell)}}^*$ represents a potential improvement in the objective if the best column for class $k$ is added. Thus, a conservative estimate for the lower bound can be derived by subtracting the maximum potential improvement from the current best objective value. We refer to the work by \cite{holmberg2003} for a similar proof. $Z_{LB}$ can be used to terminate the CG procedure when a specified level of optimality gap is attained. Theoretically, the PSP has to be exactly solved at least $K$ times (once per class) during the column generation algorithm to find an optimal solution for the master problem \eqref{eq:master_model}. 
However, solving the PSP exactly is not a good strategy for devising in a machine learning algorithm. The PSP is very difficult to solve as it is combinatorial in nature. Indeed, Proposition \ref{lem:psp} shows that our PSP is NP-hard. Therefore, the required computational effort may hamper our motivation to obtain a fast method.

\subsection{Proxy Pricing Subproblem}
\label{sec:proxyPSP}
Having observed the caveats of solving the PSP in \eqref{PSP_class_partitioned}, we devise a heuristic rule generation approach that makes use of the weighted classification structure of PSP in \eqref{PSP_class}. We coin this heuristic as \textit{proxy PSP}, which is based on growing DTs with sample weights that correspond to the dual optimal variables. After constructing the tree, we visit each leaf of the tree and check whether the resulting rule has a negative reduced cost. The following lemma shows that indeed the samples that are misclassified would receive a positive sample weight via the dual optimal solution. Hence, when training the DT, only samples that were previously misclassified by some of the rules are considered. The proof of the lemma is skipped, since it is based on complementary slackness.

\begin{lemma}\label{lem:duals}
  Suppose $v_i^{(t)}$, $i\in\CI$ are optimal solutions of the restricted master problem at iteration $t$. Then, we obtain for all $i \in \CI$ that
  \[
    v_i^{(t)} > 0 \implies \beta_i^{(t)} = 1 \hsp\text{and}\hsp \beta_i^{(t)} = 0 \implies v_i^{(t)} = 0.
  \]
\end{lemma}

\begin{figure}[h!]
    \centering
    \includegraphics[width=0.75\textwidth]{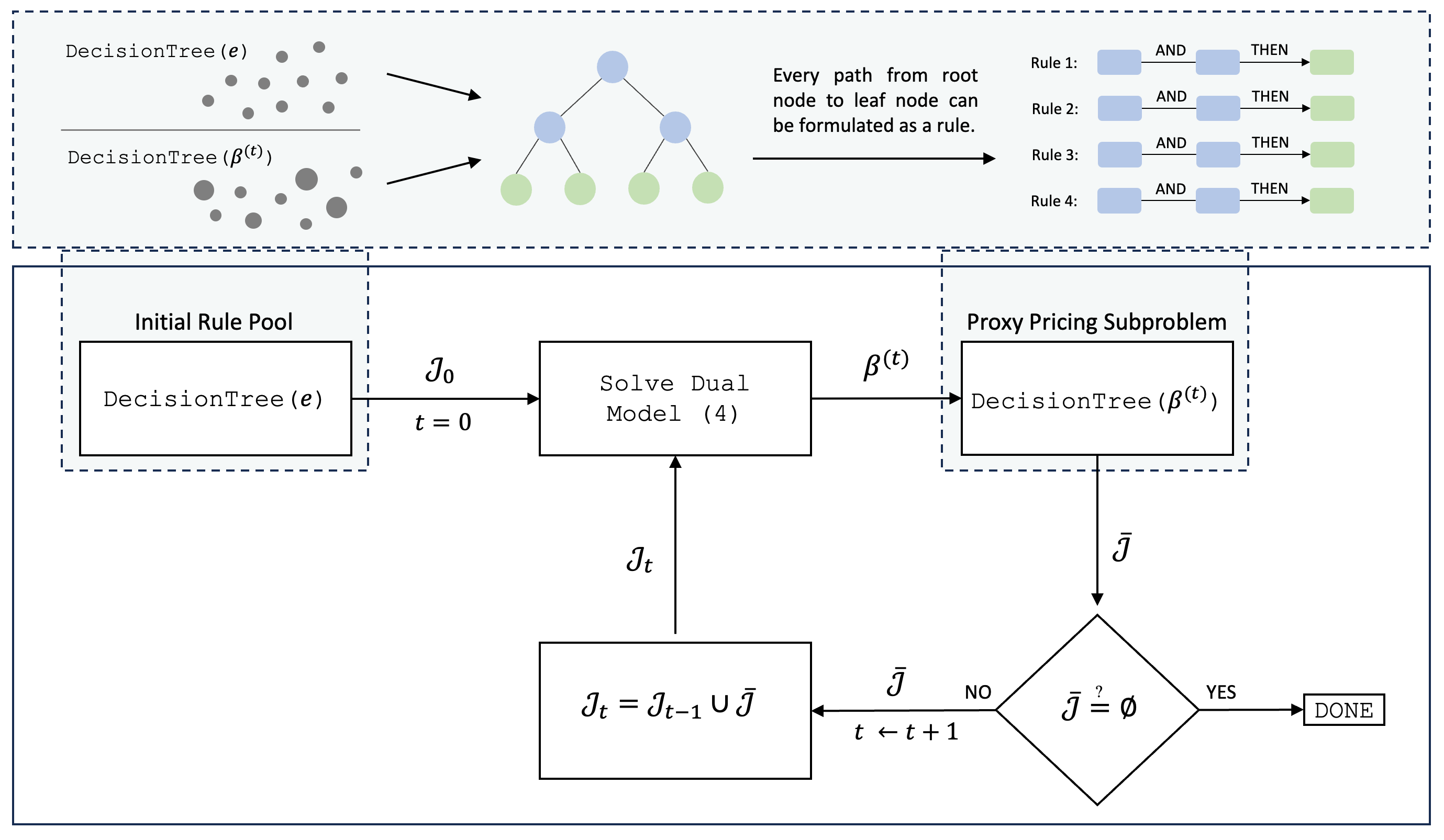}\vspace{-0.3cm}
     \caption{Proposed rule generation algorithm. The notation $\bm{e}$ stands for vector of ones.}
    \label{fig:flow_chart}  
\end{figure}

Figure \ref{fig:flow_chart} shows the proposed heuristic approach for rule generation. The procedure \texttt{DecisionTree} takes a vector of sample weights as an input and returns a set of rules $\bar{\mathcal{J}}$ (leaves of the DT) that satisfy \eqref{redcostcheck}. 
The subset $\bar{\mathcal{J}}$ thus contains those rules with negative reduced costs. Similarly, the initial rule pool, denoted by $\mathcal{J}_0$, is generated by training a DT using unit weights (denoted with \texttt{DecisionTree}(\textbf{\textit{e}})) for the samples in the upper-left part of Figure \ref{fig:flow_chart}. When we have a set of rules without such a tree structure, it might be that a sample is not covered by any rule. However, the rules obtained from a DT guarantee that every (test or training) sample is covered by exactly one rule. Thus, our CG initialization scheme is likely to prohibit having uncovered samples. In case a test sample is not covered, then we resolve this issue by assigning each uncovered sample to a default class like many rule-based models; see for instance, \cite{Furnkranz1999RuleLearning,Lawless-etal-2021FairBoolean}. Training DTs with sample weights as such boils down to solving a \textit{proxy pricing subproblem}. Fortunately, DTs with sample weights can be trained extremely fast using standard ML libraries. We note that DTs are known to overfit when grown fully, which might hamper the performance of the proposed heuristic. In practice, hyperparameter tuning through cross-validation is applied to eliminate the risk of generating a tree that is too specific on the training data. In our proposed heuristic, we further limit the \texttt{max\_depth} parameter of the trained tree to focus on interpretability. 
In addition, we associate costs for the rule length (which corresponds to the tree depth), and hence shorter rules might be preferred. This ensures that the generated rules remain interpretable and that the rules do not memorize (overfit) the training set.

The essence of our rule learning framework is its flexibility to incorporate rule costs and additional constraints while maintaining scalability. In the subsequent sections, we demonstrate two such cases where interpretability and fairness concepts can be modeled. Interpretability is addressed by taking advantage of the rule weights and their associated costs. Fairness is handled by using constraints within our LP \eqref{eq:master_model}.

\subsection{Interpretability and Fairness}
\label{section::sub::interpretability}

An interpretable learning method can be considered as global or local \citep{Molnar2022_IntML}. The former emphasizes an explanation of the underlying model, whereas the latter focuses on individual or per-sample explanations. For example, a model with a few number of rules can be good for global interpretation. In contrast, the average number of rules per sample could be used for local interpretation, \textit{i.e.}, for personalized explanations or recommendations. In this case, the total number of rules might not be as important as it is in global interpretation. In fact, local interpretability is usually overlooked by existing studies where the emphasis is given to global interpretation \citep{Lawless-etal-2021FairBoolean,DashGunlukWei2018BooleanCG,WangRudinVelezLiu2017BayesianRuleSet,WeiDashGaoGunluk2019GeneralizedLinRuleModels}. These studies implicitly assume that producing a few rules would make their approaches more interpretable. This can be correct from a global interpretation perspective. However, this also presumes that the same set of rules are to be used to explain the underlying reasoning behind every decision. Consider, for instance, explaining the rejection decisions for loan applications using the same set of rules to all individuals. Rejected people might expect individualized and specific explanations to be able to change their status in the future. A similar deduction can be made to treat patients using individualized recipes, where local interpretation becomes highly desirable. In addition, we consider the average rule length per sample as an indicator for local interpretability at the sample level.

Recall that our work is centered around local interpretability instead of global. However, our rule generation algorithm can incorporate well-known interpretability measures such as the number of rules, the length of a rule (\textit{i.e.}, number of conditions used) and the number of covered samples by the rule (see \cite{LakkarajuBachLeskovec2016InterpretableDecisionSets} for their definitions) by representing them as the costs associated with the rules at the global level. For instance, the number of conditions of a rule can be set as its cost, and then, our rule generation algorithm can be applied to choose the best possible combination of \textit{short} rules.

Accurate or interpretable predictions do not automatically guarantee fairness. Nevertheless, interpretability is usually seen as a prerequisite to ensure fairness and mitigate bias in data \citep{Arrieta2020XAI}. Fairness is divided into two as \textit{individual} and \textit{group} fairness by \cite{FairMLBook_2019}. The former emphasizes similar treatments for similar samples (individuals), and the latter prioritizes similar treatments among groups constituting sensitive attributes. In this work, we focus on group fairness.

A sensitive attribute is a feature that is considered to be subject of unfairness and may contain multiple categories. For example, gender is a sensitive attribute that could have two categories such as male and female. A protected group is defined as one of the categories of the sensitive attribute. Fairness notions are usually limited to regression and binary classification under a sensitive attribute with two protected groups. It is relatively easy to measure the differences of algorithmic decisions based on the so-called positive or negative (binary) classes between two protected groups where one is usually considered (de)favored. 

Transforming fairness discussion from a binary classification setting into a multi-class classification with multiple protected groups is not straightforward \citep{Denis-etal-2021MClassFairness,Alghamdi-etal-2022MultiClassFairness}. 
For an overview on fairness notions from the literature for both binary and multi-class classification, we refer to \ref{app:fairness}.
We introduce two notions addressing multiple classes and multiple protected groups for group fairness that are based on avoiding disparate mistreatment: \textit{Disparate Mistreatment per Class} (DMC) and \textit{Overall Disparate Mistreatment} (ODM). To our knowledge, this is the first attempt to generalize fairness notions to multiple classes and multiple protected groups addressing disparate mistreatment by using an in-processing technique \citep[see][]{Wan-etal-2023In-Processing}.

We map our fairness definitions for DMC and ODM into mathematical conditions such that they can be formulated as constraints within our LP model \eqref{eq:master_model}. Let $\mathcal{G}$ be the set of protected groups $g \in \mathcal{G}$ having distinct sensitive characteristics. 
Each sample $i\in \mathcal{I}$ is associated with a protected group label $G \in \mathcal{G}$ and a class label $y \in \mathcal{K}$. When samples $i \in \mathcal{I}$ belong to group $g \in \mathcal{G}$ and samples $j \in \mathcal{I}$ belong to group $g' \in \mathcal{G}$, and both groups of samples share the same class label, \textit{i.e.}, $y_i = y_j = k$, then the fairness criteria must hold between groups $g$ and $g'$.
 
\paragraph{Disparate Mistreatment per Class (DMC)}Disparate mistreatment requires an equal error rate among all groups $g,g' \in \mathcal{G}$ for each class $k \in \mathcal{K}$. DMC is defined as follows:
\begin{equation} \label{eq:Fcondition_dmc_1}
    P(\hat{y}\neq y |y = k, G=g) = P(\hat{y}\neq y|y = k, G=g'), \qquad k \in \CK \text{ and } g, g' \in \mathcal{G},
\end{equation}
which states that, empirically, the probability of incorrectly predicting each class $k \in \mathcal{K}$ should be equal among the protected groups. In practice, \eqref{eq:Fcondition_dmc_1} may be too strict to satisfy, and finding such a classifier under these conditions can be unrealistic. As a remedy, a certain level of unfairness $\epsilon \geq 0$ may be permitted such that the inequalities
\begin{equation} \label{condition_eq_op_2}
    \left| P(\hat{y}\neq y |y = k, G=g) - P(\hat{y}\neq y|y = k, G=g') \right| \leq \epsilon , \qquad k\in \CK \text{ and } \ g, g' \in \mathcal{G}
\end{equation}
hold, and disparate mistreatment is small. We translate the restrictions \eqref{condition_eq_op_2} into constraints that could be added to our LP model \eqref{eq:master_model}. Let $\mathcal{I}_{k,g} = \{ i \in \mathcal{I} : y_i = k, G=g\}$ be the set of indices designating the samples with class $k \in \mathcal{K}$ and group $g \in \mathcal{G}$ with cardinality $|\mathcal{I}_{k,g}|$. Then, the left-hand side of \eqref{condition_eq_op_2} for the DMC describes the difference between the classification errors of protected groups $g$ and $g'$ sharing the same class label $k$. Thus, to address group fairness using DMC, we add the following set of constraints to the LP:
\begin{align}
    & \frac{1}{|\mathcal{I}_{k,g}|}\sum_{i\in\mathcal{I}_{k,g}} v_i- \frac{1}{|\mathcal{I}_{k,g'}|} \sum_{i\in\mathcal{I}_{k,g'}} v_i \leq \epsilon, \qquad k \in \CK \text{ and } g, g' \in \mathcal{G}, \label{dmc_c1}\\
    &\frac{1}{|\mathcal{I}_{k,g'}|} \sum_{i\in\mathcal{I}_{k,g'}} v_i- \frac{1}{|\mathcal{I}_{k,g}|} \sum_{i\in\mathcal{I}_{k,g}} v_i \leq \epsilon, \qquad k \in \CK \text{ and } g, g' \in \mathcal{G}. \label{dmc_c2}
\end{align}

The first term in \eqref{dmc_c1} stands for the fraction of misclassification error made over the samples belonging to the protected group $g$ for class $k$. Likewise, the second term is for the protected group $g'$. Due to the absolute value function in \eqref{condition_eq_op_2}, we reverse the left-hand side of \eqref{dmc_c1} in \eqref{dmc_c2}. Recall that $v_i$ values in our master problem \eqref{eq:master_model} measure an approximate classification error rather than counting the number of misclassifications. Therefore, \eqref{dmc_c1} and \eqref{dmc_c2} can be considered as soft constraints yielding approximate percentages of unfairness among protected groups and classes in DMC.

When the constraints based on DMC are defined for binary classification and for only two protected groups, that is, $|\mathcal{K}|=2$ and $|\mathcal{G}| = 2$, \eqref{dmc_c1} and \eqref{dmc_c2} reduce to the fairness definition of \textit{Equalized Odds} used by \cite{Lawless-etal-2021FairBoolean}. Equalized odds enforces equal false positive rate and false negative rate for both protected groups (see \ref{app:fairness}). A relaxed version of Equalized Odds is the definition of \textit{Equal Opportunity} that merely mandates an equal false negative rate between two protected groups. This can be achieved by constructing constraints \eqref{dmc_c1} and \eqref{dmc_c2} only for the positive class. Hence, the constraints based on DMC can be used to compute the fairness scores for Equal Opportunity and Equalized Odds. Briefly, the DMC is more general than Equalized Odds, since the terms false positive and false negative in Equalized Odds vanish for multi-class classification.  
 
\paragraph{Overall Disparate Mistreatment (ODM)} ODM is a generalization of disparate mistreatment (DM) \citep{Zafar-etal-2019FairnessFlex}, where misclassification rates are equalized in binary classification, to multiple protected groups. Unlike DMC, ODM ignores classes and directly focuses on protected groups to equalize overall misclassification among them. 
ODM requires that among all protected groups, the overall mistreatment should be the same:
\begin{equation} \label{eq:Fcondition_odm_1}
    P(\hat{y}\neq y| G=g) =  P(\hat{y}\neq y| G=g'), \qquad g, g' \in \mathcal{G},
\end{equation}
where \eqref{eq:Fcondition_odm_1} indicates that the probability of incorrect predictions should be equal among protected groups. As in DMC, a certain threshold of $\epsilon \geq 0$ is used to derive the ODM constraints
\begin{equation} \label{eq:condition_total}
    \left| P(\hat{y}\neq y | G=g) - P(\hat{y}\neq y | G=g') \right| \leq \epsilon, \qquad g, g' \in \mathcal{G}.
\end{equation}
 
Recall that, $ P(\hat{y} \neq y| G=g)$, describes the total classification error of all samples in the protected group, and the left-hand side of \eqref{eq:condition_total} describes the difference of mistreatment between two different protected groups. Let $\mathcal{I}_g = \{ i\in \mathcal{I}: G=g\}$ be the set of samples in protected group $g$ and $|\mathcal{I}_g|$ represent its cardinality. Then, we map \eqref{eq:condition_total} into the following set of constraints and add them to our LP model \eqref{eq:master_model}:
\begin{align}
    & \frac{1}{|\mathcal{I}_g|} \sum_{i\in\CI_g} v_i-\frac{1}{|\mathcal{I}_{g'}|} \sum_{i\in\CI_{g'}} v_i \leq \epsilon, \qquad g, g' \in \mathcal{G}, \label{eq:constraint_odm1}\\ 
    & \frac{1}{|\mathcal{I}_{g'}|} \sum_{i\in\CI_{g'}} v_i- \frac{1}{|\mathcal{I}_g|} \sum_{i\in\CI_g}v_i \leq \epsilon, \qquad g, g' \in \mathcal{G}. \label{eq:constraint_odm2}
\end{align}
The first and the second terms in \eqref{eq:constraint_odm1} stand for the fraction misclassification error made over the samples belonging to the protected groups $g$ and $g'$, respectively. Note that these terms accumulate the error made for all classes $k \in \mathcal{K}$.
 
For the sake of brevity, we skip the details on the derivation of the dual problems and the resulting calculations of the column generation algorithm for the cases when the fairness constraints corresponding to DMC and ODM are added. It is important to note that the resulting PSP is exactly the same as in \eqref{PSP_class}. Nevertheless, new dual variables are defined for the corresponding constraints \eqref{dmc_c1}--\eqref{dmc_c2} for DMC, and \eqref{eq:constraint_odm1}--\eqref{eq:constraint_odm2} for ODM, respectively, to address fairness. This affects only the dual variable values $\bm{\beta}$ obtained from the resulting dual formulations as the second constraint set in \eqref{RMP_d_class} takes a slightly different form without changing the calculation of the reduced cost values in \eqref{redcostcheck}. Other aspects of the column generation framework remain the same as before. Besides, the addition of fairness constraints does not affect the feasibility of the dual problem nor requires another initialization scheme of the columns. As a remark, a similar conclusion directly follows for interpretability, as we are already using a penalization strategy and have no additional constraints with respect to interpretability.

\section{Numerical Experiments}\label{sec:num_exp}

We evaluate the performance of the proposed \textit{rule generation algorithm} (RUG) on a collection of standard datasets for classification with the sample sizes ranging from 178 to 245057, the number of classes from 2 to 11, and the number of features from 3 to 166 (see \ref{app:exp} for details). 
We first compare the exact solution of the PSP formulation \eqref{PSP_explicit2} against the results of the proxy PSP in Section \ref{sub:proxy_exact}.
Then, we compare our results against four recent algorithms based on mathematical optimization that are the most related ones to our methodology and prioritise not only classification performance but also interpretability (and fairness). These are fast sparse DT optimization (FSDT) by \cite{McTavish-etal-2021}, binary optimal classification trees (BinOCT) by \cite{VerwerZhang2019Learning}, a heuristic for learning DTs (DT-h) by \cite{PATEL2024106579} and rule sets via column generation (CG and FairCG for fairness) by \cite{Lawless-etal-2021FairBoolean}. FSDT, BinOCT and DT-h are tree-based methods, and CG is a rule-based method. All these approaches consider interpretability. 
The data must be in binary form for these benchmark methods, specifically FSDT, BinOCT, and CG, and thus, they require pre-processing of the datasets. 
We use the same binarization scheme as described in the reference study \cite{Lawless-etal-2021FairBoolean} for CG, FairCG, FSDT, and BinOCT. To that end, a binary variable is introduced for each category (and for its negation) of a feature. Numerical features are binarized so that the quantiles of the dataset are used as the splitting values and there exists a decile of data between every split. We use ten quantiles for splitting the data as suggested by \cite{Lawless-etal-2021FairBoolean}. Then, for every split value $X$, two binary variables are introduced, one for $x_p \leq X$ and the other for $x_p > X$ where $x_p$ is feature $p$ with numerical values. 

In all our experiments, we resort to a five-fold cross-validation scheme combined with grid search for hyperparameter tuning, and then fit the model with the best parameters on the complete train set. Subsequently, we apply the trained model to a previously unseen test set. All scores reported in this section are obtained on this test set. For FSDT, BinOCT and DT-h, we select \texttt{max\_depth} parameter from the set \{3,5,10\} for maximum tree depth. For RUG, we chose this value from the set \{3,5\} to generate rules using DTs. Furthermore, for RUG we select the penalty parameter $\lambda$ (\texttt{pen\_par}) from the set \{0.1,1.0,10.0\} and the iteration limit (\texttt{max\_RMP\_calls}) from \{5, 10, 15\}. We should note that the convergence of a classical column generation procedure might take too long in practice, and here, our goal is to achieve an effective balance between accuracy, interpretability, and computational efficiency for RUG. By limiting \texttt{max\_RMP\_calls}, this process does not necessarily need to achieve convergence of the CG procedure in the strict mathematical sense. However, observe that there are multiple rules corresponding to each DT and all the rules with negative reduced cost are added to the master problem at each iteration. Therefore, depending on the value of the \texttt{max\_depth} parameter, it is possible to find tens of rules with a negative reduced cost per iteration. We often observed that the improvement in accuracy that comes with more iterations becomes negligible and that this characteristic of our heuristic approach, \textit{i.e.}, adding multiple columns per iteration, also accelerates the column generation procedure of the RUG. We select the rules whose weight $w_j$ exceeds 0.05 for prediction. This helps us to reduce the number of rules by discarding the negligible ones. Lastly, we assign the number of features as the cost coefficient of a rule in RUG. For CG, we also perform a grid search for the so-called \textit{complexity parameter} $C$ defined by \cite{Lawless-etal-2021FairBoolean} as an indicator of sparsity. The authors report results for mean values of $C$ per dataset. We cross-validate the complexity parameter $C$, considering a set of values around its reported mean value. For example, for a reported mean $C$ of 25.0, we choose values 20.0, 25.0, and 30.0 for the grid search. The grid search set is limited to three values, all multiples of five, to ensure computational tractability. We impose a running time limit of 300 seconds for all methods. Lastly, all our programs are implemented in Python version 3.10.2 on a computer with an Apple-M1-Pro processor. The resulting LPs and MIP formulations are solved using the Gurobi 10.0.2 solver under its default settings \citep{gurobi}. To reproduce the numerical experiments we refer the reader to our repository\footnote{\href{https://github.com/sibirbil/RuleDiscovery}{https://github.com/sibirbil/RuleDiscovery}}.

\subsection{Proxy PSP vs. Exact PSP}\label{sub:proxy_exact}

\begin{figure}[h]
    \centering
    \includegraphics[width=0.75\textwidth]{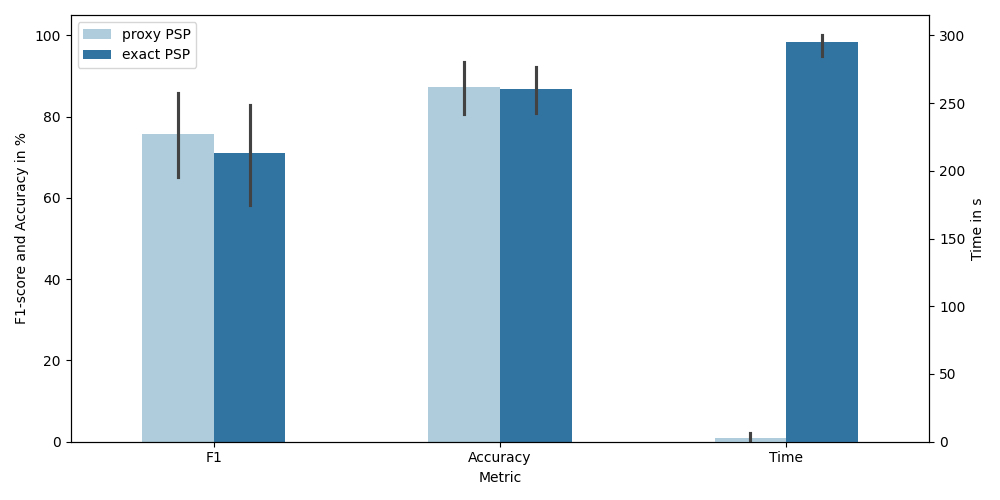}
    \caption{Comparison of metrics between RUG with proxy PSP and RUG with the exact solution of the PSP on the test set.}
    \label{fig:proxy_exact}
\end{figure}

We compare the efficiency and performance of RUG with the optimal solutions of the PSP formulation (\ref{PSP_explicit2}) against RUG with the proxy PSP on binary classification problems from \ref{app:dataset_properties}. As the PSP formulation \eqref{PSP_explicit2} requires the data to be in binary form, we apply the process described above to discretize the features. Due to the computational intensity of the PSP formulation, the cross-validation scheme with grid search for hyperparameter tuning is not sensible as it requires an excessively long time. Hence, we apply this process only to the heuristic version of RUG with the proxy PSP, and then solve the PSP formulation only once with the parameters that worked best for the heuristic approach. In Table \ref{tab:results-rug-exact-short} in \ref{app:proxy_exact}, we report the computation time in seconds in parentheses, the F1-score, and the accuracy. 
When data is imbalanced, relying only on the accuracy measure can be specious. Therefore, we have also used the F1-score that is less conservative than the accuracy for imbalanced datasets \citep{Japkowicz2013_AssessmentMetrics}. Figure \ref{fig:proxy_exact} shows the average performance of both approaches over all datasets, \textit{i.e.,} the averaged scores of Table \ref{tab:results-rug-exact-short}.

For each problem, we set a time limit of 300s, after which the current iteration was finished, and then training was stopped and results were inspected. This implies that the number of iterations set by the parameter \texttt{max\_RMP\_calls} may not be reached, and hence, the resulting model has few rules and performance may not be optimal. The heuristic works very fast and training of the model completes within a couple of (milli)seconds, however RUG with the exact solution of the PSP formulation is computationally quite expensive and almost always hits the imposed time limit. Both the F1-score and accuracy show that the two versions of RUG perform similarly well on most of the datasets. 
On average, the heuristic performs better by about 5\% and 1\% in terms of F1-score and accuracy, respectively. Possibly, the heuristic performs on average slightly better on the test sets than the exact formulation as we fit \textit{simple}, \textit{i.e.}, shallow, DTs that avoid overfitting. Furthermore, recall that we did not perform cross validation for the RUG with the exact solution of the PSP formulation due to its high computational demand. Instead, we obtained results just once using the parameters that worked best for the heuristic version. It may be that the RUG with the exact solution of the PSP formulation would yield other results with different parameters. This could explain why the heuristic version may occasionally outperform the exact solution of the PSP formulation. Overall, the heuristic proves to be a good alternative to the exact approach as it reaches very good and similar performance scores within a fraction of the time.

In the remainder of this section we will compare RUG to other algorithms from the literature. For these comparisons we will rely on the heuristic approach, which we will simply refer to as `RUG'.

\subsection{Experiments for Interpretability}
\label{sub:exp_interpretability}

We first compare the performance of our algorithm in terms of interpretability against those of FSDT, BinOCT, DT-h, and CG. Table \ref{tab:average_interpretability} provides a summary of our results over the binary classification datasets from \ref{app:dataset_properties}. To be specific, each cell in Table \ref{tab:average_interpretability} is calculated by taking the average of the corresponding columns in Table \ref{tab:results} in \ref{app:interpretability}. The first column represents the used method. We report computation time in seconds for each method, along with the F1-score and accuracy as performance indicators under the columns entitled ``CPU (s)'', ``F1 (\%)'', and ``Acc. (\%)'', respectively. We present the number of rules (``NoR'') and average rule length (``Avg. RL'') in the fifth and the sixth columns to evaluate global interpretability performances. For the tree-based methods (FSDT, BinOCT, and DT-h), the number of leaf nodes in their DTs is reckoned as the number of rules. For RUG, the ``NoR'' corresponds to the number of rules having a weight satisfying the threshold, \textit{i.e.}, $w_j \geq 0.05$. For CG, NoR is the number of rules generated by excluding the default rule of assigning an uncovered sample to the majority class from the calculations as this is not generated by the CG. Then, the ``Avg. RL'' is the average number of levels in the leaf nodes of the DTs for the tree-based methods FSDT, BinOCT, and DT-h, whereas, for RUG and CG, this value is the number of conditions (literals) used in the rules. The last two columns stand for the average number of rules per sample (``Avg. NoRpS'') and the average rule length per sample (``Avg. RLpS'') as an indicator of local interpretability. The average number of rules per sample, ``Avg. NoRpS'', gives the number of rules that contribute to classification per sample, averaged over all samples of the test set. The ``Avg. NoRpS'' is one for the tree-based methods as every sample belongs to a non-overlapping partition of the feature space, resulting in a single rule. In contrast, the rule-based methods RUG and CG can offer multiple rules per sample. This implies that RUG and CG do not offer the same rule (\textit{i.e.}, recipe) for an individual which is advantageous when we prioritize local interpretability. Clearly, there is a trade-off, and the ``Avg. NoRpS'' should not be excessive as well. RUG strikes a good balance in this respect, offering informative explanations without excessive complexity. The ``Avg. RLpS'' is similar to the average rule length but takes the average of rule length per sample. Briefly, shorter values of ``Avg. RLpS'' is considered better for local interpretability.

{\renewcommand{\arraystretch}{0.7}
\captionsetup[table]{labelfont=small, textfont=small}
\begin{table}[ht]
\centering
\caption{Average results on the test set of RUG, FSDT, BinOCT, and CG over binary classification datasets.}
\label{tab:average_interpretability}
\resizebox{0.8\textwidth}{!}{%
\begin{tabular}{@{}lrrrrrrr@{}}
\hline \hline
                                       & \textbf{CPU (s)} & \textbf{F1 (\%)}  & \textbf{Acc. (\%)} & \textbf{NoR} & \textbf{Avg.RL} & \textbf{Avg.NoRpS} & \textbf{Avg.RLpS} \\ \midrule
 RUG & 3.78 &	76.96 &	88.15 &	42.11 &	2.96 & 4.11 & 3.01 \\
 FSDT & 163.23 & 74.14 & 86.63 & 18.61 & 4.27 & 1.00 &	3.83\\
  BinOCT     & 300.00 &	35.27	& 64.65	& 8.00 & 3.00 &	1.00 &	3.00 \\
 CG & 283.38 & 68.70 & 85.59 & 2.89 & 4.00 &	1.21 & 3.92 \\
 DT-h & 164.67	& 78.45 & 87.71 & 579.11 & 7.44 & 1.00 & 7.44 \\
 \hline \hline
\end{tabular}%
}
\end{table}}

The reported results in Table \ref{tab:results} of \ref{app:interpretability} are obtained by each model applied to a hold-out test set, using the parameters deemed best through cross-validation on the training set. Overall, RUG outperforms the other methods by more than an order of magnitude in terms of CPU times. RUG also demonstrates superior performance in terms accuracy compared to the other methods.
In fact, RUG achieves an average accuracy of 88.15\%, followed by 87.71\% by DT-h and 86.63\% by FSDT. Regarding the F1-score, RUG is the second best performing model with a score of 76.96\% shortly behind DT-h (78.45\%). Interestingly, DT-h actually performs the same or worse than RUG on 12 out of the 18 datasets, however performs substantially better on some datasets which leads to a slightly better average score. Overall, while DT-h performs quite well, the number of rules is very large, with an average of 579.11 rules, by that rendering the model non-interpretable. RUG has an average of 42.11 rules, which exceeds those of the remaining models, however shows that a high accuracy can be reached with substantially less rules than DT-h. Furthermore, the ``Avg. RL'' and the ``Avg. RLpS'' for RUG are comparable to those of other methods, indicating that RUG achieves enhanced performance while preserving local interpretability. ``Avg. NoRpS'' for RUG is higher than the other methods, owing to the fact that in a tree-based structure like FSDT, BinOCT, and DT-h each sample gets an assignment by only one rule. Nonetheless, RUG manages to find a fine line for local interpretability by avoiding both excessively high and noticeably low numbers of rules per sample. 

We also notice that RUG outperforms the other methods in the case of imbalanced datasets. For example, all methods perform comparably well on \textsc{mammography} in terms of accuracy, with values ranging from 97.99\% (BinOCT) to 98.84\% (RUG). However, looking at the F1-score, we observe that, with a score of 70.45\%, RUG outperforms all other methods except for DT-h by almost 20\%. We can see a similar trend for other imbalanced datasets, such as \textsc{musk} and \textsc{oilspill}.

Our comparison with other MOBAs is limited to binary classification for FSDT, BinOCT, and CG as they do not address multi-class classification. A common workaround is to apply a \textit{one-vs-rest} scheme, where a classifier for each class is build and the prediction is derived using the predicted probabilities for each class returned by each of the classifiers. Unfortunately this procedure is also not straightforward for the discussed MOBA methods as it requires a predicted probability per class, which these methods do not provide.
RUG can directly handle multiple classes without resorting to a \textit{one-versus-rest} scheme. In Table \ref{tab:results_multiclass}, we present results obtained with RUG on a set of multi-class classification problems. We used the same setup for cross-validation and hyperparameter tuning as described above. Although the number of rules reaches to 83.00 for the \textsc{sensorless} dataset, the average rule length, average number of rules per sample, as well as average rule length per sample are relatively small for all datasets. We also report results obtained with DT-h, again using the same set-up for hyperparameter tuning as reported previously. Additionally, to include more models to compare to, we report results obtained with traditional machine learning models, specifically DT, random forest (RF), AdaBoost (ADA) \citep{FreundShapire1997RealADAboost}, and light gradient-boosting machine (LightGBM) \citep{LightGBM}. In general RUG reaches a similar performance in terms of classification accuracy when compared to the other models, in most cases with substantially less rules especially when compared to RF, ADA, and LightGBM. For most datasets RUG performs slightly worse than a DT, however with slightly less (``NoR'') and also shorter (``Avg. RL'' and ``Avg. RLpS'') rules, which makes it more interpretable when considering global interpretability. Naturally, to classify single sample, a DT utilizes only one rule, whereas RUG uses a subset of the rules generated. This may still be beneficial for local interpretability as the subset of rules gives a more personalized and specific insight into the classification. The comparison to DT-h is less straightfoward as it seems to depend on the dataset. For example, RUG (with 7 rules) performs a bit worse than DT-h (32 rules) on \textsc{ecoli}, whereas the pattern is reversed for \textsc{glass}, where RUG (17 rules) outperforms DT-h (8 rules). This seems intuitive as more rules may lead to higher accuracy. However, on \textsc{seeds}, RUG reaches the same high performance as DT-h with only 10 rules, while DT-h consists of 1024 rules. Overall, RUG seems to perform very well and can reach competitive scores when compared against the other models. For the \textsc{wine} dataset RUG even reaches the same performance as a powerful LightGBM model, with less computation time, shorter and substantially less rules (RUG returns 14 rules, while LightGBM has 1573 rules).

{\renewcommand{\arraystretch}{0.8}
\captionsetup[table]{labelfont=small, textfont=small}
\begin{table}[ht]
\centering
\caption{Comparison of the performance of RUG to DT-h and traditional machine learning models on multi-class datasets using five-fold cross validation and grid search for hyperparameter tuning on a hold-out test set.}
\label{tab:results_multiclass}
\resizebox{\textwidth}{!}{%
\begin{tabular}{@{}p{0.19\textwidth}p{0.1\textwidth}R{0.13\textwidth}R{0.13\textwidth}R{0.13\textwidth}R{0.15\textwidth}R{0.1\textwidth}R{0.12\textwidth}R{0.1\textwidth}@{}}
\hline \hline
                                       & & \textbf{CPU (s)} & \textbf{weighted F1 (\%)}  & \textbf{Acc. (\%)} & \textbf{NoR} & \textbf{Avg. RL} & \textbf{Avg. NoRpS} & \textbf{Avg. RLpS} \\ \midrule
 \multirow{6}{*}{\textsc{ECOLI}} & RUG & 0.03 & 75.54 & 77.94 & 7.00 & 2.29 & 1.29 & 2.00 \\
 &  DT-h &   49.99 &   80.83 &   82.35 &   32.00 &   5.00 &   1.00 &   5.00 \\
 \cdashline{2-9}
 & DT & 0.01 & 78.36 & 79.41 & 8.00 & 3.00 & 1.00 & 3.00 \\
 & RF & 0.10 & 88.16 & 88.24 & 5060.00 & 5.71 & 150.00 & 5.61 \\
 & ADA & 0.07 & 53.91 & 66.18 & 200.00 & 1.00 & 100.00 & 1.00 \\
 & LightGBM & 1.29 & 87.30 & 88.24 & 14097.00 & 4.10 & 2400.00 & 2.31 \\
 \hline
 \multirow{5}{*}{\textsc{GLASS}} & RUG &  0.03 & 60.57 & 62.79 & 17.00 & 2.47 & 2.35 & 2.50 \\
 & DT-h &   7.49 &   52.70 &   55.81 &   8.00 &   3.00 &   1.00 &   3.00 \\
 \cdashline{2-9}
 & DT & 0.01 & 63.84 & 62.79 & 32.00 & 5.81 & 1.0 & 5.79 \\ 
& RF & 0.07 & 77.42 & 76.74 & 2717.00 & 5.51 & 100.00 & 5.57 \\
& ADA & 0.07 & 24.34 & 32.56 & 200.00 & 1.00 & 100.00 & 1.00 \\
& LightGBM & 0.65 & 81.57 & 81.40 & 7677.00 & 3.49 & 1200.00 & 3.08 \\
 \hline
\multirow{5}{*}{\textsc{SEEDS}} & RUG & 0.03 & 90.43 & 90.48 & 10.00 & 2.60 & 2.40 & 3.50 \\
& DT-h* &   40.17 &   90.41 &   90.48 &   1024.00 &   10.00 &   1.00 &   10.00 \\
\cdashline{2-9}
& DT & 0.01 & 92.77 & 92.86 & 11.00 & 3.91 & 1.00 & 3.31 \\
& RF & 0.12 & 90.43 & 90.48 & 2337.00 & 3.92 & 200.00 & 3.37 \\
& ADA & 0.06 & 65.00 & 71.43 & 200.00 & 1.00 & 100.00 & 1.00 \\
& LightGBM & 0.17 & 92.77 & 92.86 & 1699.00 & 2.63 & 300.00 & 2.73 \\
 \hline
 \multirow{5}{*}{\textsc{SENSORLESS}} & RUG & 11.39 & 79.18 & 78.42 & 83.00 & 4.25 & 5.33 & 4.38 \\
 & DT-h* & 194.49 & 66.69 & 71.15 & 32.00 & 5.00 & 1.00 & 5.00 \\
 \cdashline{2-9}
 & DT & 2.06 & 98.28 & 98.28 & 425.00 & 11.69 & 1.00 & 11.42 \\
 & RF & 17.56 & 98.44 & 98.44 & 56664.00 & 12.10 & 100.00 & 11.06 \\
 & ADA & 33.30 & 32.54 & 45.36 & 300.00 & 1.00 & 150.00 & 1.00 \\
 & LightGBM & 42.52 & 99.97 & 99.97 & 48414.00 & 4.54 & 3300.00 & 3.38 \\
 \hline
 \multirow{6}{*}{\textsc{WINE}} & RUG & 0.05 & 97.24 & 97.22 & 14.00 & 1.57 & 4.39 & 1.61 \\ 
 & DT-h &   79.65 &   97.24 &   97.22 &   8.00 &   3.00 &   1.00 &   3.00 \\
 \cdashline{2-9}
  & DT & 0.01 & 94.45 & 94.44 & 5.00 & 2.60 & 1.00 & 2.44 \\
  & RF & 0.06 & 97.24  & 97.22 & 639.00 & 2.74  & 100.00 & 2.73 \\
  & ADA & 0.07 & 88.49 & 88.89 & 200.00 & 1.00 & 100.00 & 1.00 \\
  & LightGBM & 0.19 & 97.24 & 97.22 & 1573.00 & 2.51 & 300.00 & 2.62 \\
 \hline \hline
\multicolumn{9}{l}{\footnotesize Hyperparameter sets for RUG --  max\_depth = \{3,5\}, pen\_par = \{0.1,1.0,10.0\}, max\_RMP\_calls = \{5,15,30\}. For DT-h -- max\_depth = \{3,5,10\}. Time limit 300s.} \\
\multicolumn{9}{l}{\footnotesize Hyperparameter sets for DT -- max\_depth = \{3,5,7,9,11,13,15\}; for ADA -- n\_estimators = \{100,150,200,250,300\}; for RF and GB -- max\_depth =} \\
\multicolumn{9}{l}{\footnotesize \{3,5,7,9,11,13,15\}, n\_estimators = \{100,150,200,250,300\}; * hyperparameter sets excluded some parameters in cases where those lead to execution failure} 
\end{tabular}%
}
\end{table}}

\subsection{Experiments for Fairness}
\label{sub:fairness_results}

We analyze the performance of RUG with fairness constraints, coined as FairRUG, and compare it to FairCG, which is shown to be superior to several state-of-the-art methods by \cite{Lawless-etal-2021FairBoolean}. As FairCG is designed for binary classification and one sensitive attribute with two groups, our comparison is performed on such datasets. Subsequently, we shall also provide our results for multi-class problems and/or datasets with a sensitive attribute with several groups. The details of the fairness datasets including the sensitive attributes used and their associated protected groups are reported in Table \ref{table:datasets-fairness} of \ref{app:dataset_properties}.

Table \ref{tab:fairness_binary} presents the results of FairRUG and FairCG that are subject to two fairness constraints concerning \textit{Equal Opportunity} (EOP) and DMC. Observe that FairCG does not address ODM and thus we do not use it for comparison. The first two columns represent the name of the dataset and the method, followed by the computation time in seconds, the F1-score, and the accuracy as before. Columns 6 and 10 provide the fairness scores in percentages under ``DMC (\%)'' and ``EOP (\%)''. The EOP (see \ref{app:fairness} for definition) is obtained by relaxing the DMC constraints \eqref{dmc_c1} and \eqref{dmc_c2} for the negative class in FairRUG. The DMC generalizes the fairness definition \textit{Equalized Odds} (EOD) introduced for binary classification. FairCG handles \textit{Equalized Odds} (EOD) which is a special case of DMC. In particular, the DMC and EOD are equivalent to each other and they both measure the difference between false negatives and the difference between false positives of two protected groups in binary classification problems. In short, we simply used DMC for consistency and clarity. 

We use five-fold cross-validation for hyperparameter tuning as before. For FairRUG and FairCG, we mostly use the same setup as described for the RUG and CG. However, we set the rule weight threshold of FairRUG to zero and set the rule costs to one since our focus is now on fairness rather than interpretability. Fairness constraints \eqref{dmc_c1}, \eqref{dmc_c2}, \eqref{eq:constraint_odm1} and \eqref{eq:constraint_odm2} bring in an additional hyperparameter $\epsilon$ that needs fine-tuning, which caps the unfairness. Hence, we select $\epsilon$ from the set \{0, 0.01, 0.025, 0.05\} for FairRUG. In their study, \citet{Lawless-etal-2021FairBoolean} state that the best fairness outcomes are obtained with a cap of $\epsilon = 0.025$, and thus, we set $\epsilon = 0.025$ for the FairCG to avoid excessive computational requirements.
As usual, we select the hyperparameters in cross-validation, however in this case based on fairness performance rather than accuracy score.

In terms of fairness, we find that FairRUG outperforms FairCG on average. In particular, FairRUG yields the highest fairness scores under DMC constraints for two out of the three datasets. Considering the EOP, FairRUG achieves higher scores for all three datasets. 
We further see that FairRUG achieves a higher predictive performance, characterized by higher accuracy and F1-scores in all cases except for \textsc{compas} under EOP constraints. Besides, the computation times for FairRUG are significantly better than those of FairCG. In short, FairRUG is capable to strike a favorable balance between fairness and practicality.

{\renewcommand{\arraystretch}{0.9}
\captionsetup[table]{labelfont=small, textfont=small}
\begin{table}[!ht]
\caption{\centering Accuracy and fairness results on the test set of FairRUG and FairCG on binary classification problems with two protected groups.}
\label{tab:fairness_binary} 
\centering
\setlength{\tabcolsep}{8pt}
\resizebox{1\textwidth}{!}{
\begin{tabular}{@{}llrrrr|rrrr@{}}
\hline \hline
                                    &         & \textbf{CPU (s)} & \textbf{F1 (\%)} & \textbf{Acc. (\%)}   & \textbf{DMC (\%)} & \textbf{CPU (s)} & \textbf{F1 (\%)} & \textbf{Acc. (\%)} & \textbf{EOP (\%)} \\ 
\hline
\multirow{2}{*}{\textsc{ADULT}} & FairRUG & 32.98 & 63.81 & 85.08 & 89.10 & 17.37 & 64.73 & 85.83 & 92.39 \\
& FairCG & 300.00 & 11.09 & 76.86 & 93.22 & 300.00 & 47.66 & 81.45 & 64.90 \\
\hline
\multirow{2}{*}{\textsc{COMPAS}} & FairRUG & 0.04 & 68.19 & 68.47  & 51.89 & 0.77  & 63.77 & 65.62  & 98.81  \\
& FairCG &  300.00 & 47.73 & 62.36 & 17.74 & 300.00 & 59.42 & 66.05 & 63.93 \\
\hline
\multirow{2}{*}{\textsc{DEFAULT}} & FairRUG & 9.11 & 48.27 & 82.57 & 97.88 & 9.10 & 47.65 & 82.53 & 98.69 \\
& FairCG & 300.00 & 41.80 & 82.08 & 87.73 & 300.00 & 44.82 & 82.43 & 89.40 \\
\hline\hline
\end{tabular}
}
\end{table}}

Finally, we compare RUG with FairRUG under fairness constraints using DMC and ODM for the datasets with multi-class targets and/or sensitive attributes with more than two groups in Table \ref{table:fairness-multiclass}. We report the computation times in seconds, weighted F1-score, accuracy, and fairness scores for DMC and ODM. Here, the weighted F1-score is calculated using a weighted sum of the F1-scores of each class considering the number of samples from the corresponding class. For RUG no fairness constraints are required and the fairness scores are computed based on the test sample predictions after classification. Therefore, the CPU (s), weighted F1 (\%) and Acc. (\%) values are the same for both DMC (\%) and ODM (\%) for RUG. For FairRUG, fairness constraints are added separately for DMC and ODM, and hence, we present values for both models. 
In general, FairRUG is able to mitigate unfairness in the predictions made by RUG by adding fairness constraints which lead to improved fairness scores. This is expected especially when there exists a bias against a protected group in the dataset. However, RUG also shows a very high performance that equals that of FairRUG in case of both the \textsc{nursery} and \textsc{law} datasets.
First, a lack of bias in a dataset can positively impact machine learning methods that are equity-unaware but accurate like RUG. 
Second, incorporating fairness constraints may not yield equivalent benefits on the test samples due to learning being performed on the training set. Third, recall that we do not measure misclassifications per se, but instead use auxiliary variables $v_i$ to approximate classification error. Therefore, as detailed above, the percentages of fairness enforced by the constraints are only approximate. Furthermore, the scores we report here are computed afterwards on the test set to reflect the actual usage objectively. Hence, performance of RUG may be just as good or even better compared to FairRUG in some cases. We observe this property for the fairness scores on the \textsc{student} dataset, where fairness slightly deteriorates for FairRUG.

{\renewcommand{\arraystretch}{0.9}
\captionsetup[table]{labelfont=small, textfont=small}
\begin{table}[!ht]
\caption{Accuracy and fairness results on the test set of RUG and FairRUG on multiclass problems and/or with multiple protected groups.}
\label{table:fairness-multiclass} 
\centering
\setlength{\tabcolsep}{8pt}
\resizebox{1\textwidth}{!}{
\begin{tabular}{@{}p{0.15\textwidth}p{0.08\textwidth}R{0.11\textwidth}R{0.13\textwidth}R{0.1\textwidth}R{0.12\textwidth}|R{0.1\textwidth}R{0.12\textwidth}R{0.1\textwidth}R{0.1\textwidth}@{}}
\hline \hline
                                    &         & \textbf{CPU (s)} & \textbf{weighted F1 (\%)} & \textbf{Acc. (\%)}   & \textbf{DMC (\%)} & \textbf{CPU (s)} & \textbf{weighted F1 (\%)} & \textbf{Acc. (\%)} & \textbf{ODM (\%)} \\ 
\hline
\multirow{2}{*}{\textsc{ATTRITION}} & RUG & 0.15 & 30.51 & 86.05 & 97.46 & -- & -- & -- & 72.04 \\
& FairRUG & 0.23  & 25.45 & 86.05 & 99.58 & 0.48 & 34.78 & 84.69 & 78.87 \\
\hline
\multirow{2}{*}{\textsc{NURSERY}}   & RUG   & 0.75 & 100.00 & 100.00 & 100.00 & -- & -- & -- & 100.00 \\
& FairRUG & 1.11 & 100.00 & 100.00 & 100.00 & 1.00 & 100.00 & 100.00 & 100.00 \\
\hline
\multirow{2}{*}{\textsc{STUDENT}}   & RUG & 0.08 & 75.04 & 75.38 & 89.22 & -- & -- & -- & 88.26\\
& FairRUG & 0.29 & 61.15 & 60.77 & 81.54 & 0.13 & 70.31 & 70.77 & 84.96 \\
\hline
\multirow{2}{*}{\textsc{LAW}}  & RUG  & 0.44 & 100.00 & 100.00 & 100.00 & -- & -- & -- & 100.00 \\
& FairRUG & 0.69 & 100.00 & 100.00 & 100.00 & 4.32 & 100.00 & 100.00 & 100.00 \\
\hline\hline
\multicolumn{10}{l}{\footnotesize \textsc{ATTRITION} is a binary classification problem with a sensitive attribute with more than two groups. \textsc{NURSERY}, \textsc{STUDENT}, and \textsc{LAW} are multiclass problems.} \\
\end{tabular}
}
\end{table}}

While we focus our fairness discussion on computational aspects, it is important to acknowledge that concrete implementations may demand a more nuanced perspective, as a trade-off between accuracy and fairness often exists \citep[\textit{e.g.,}][]{liu2022accuracy, fuMS22}. Moreover, a key consideration is how policies on fairness influence stakeholder strategies. For instance, \cite{fuMS22} examine how firms' strategic behavior under equal opportunity requirements may not necessarily result in more fair outcomes.

\subsection{Case Study: Credit Risk Scoring}\label{sec:case_interpretability}

In this case study, we look at credit risk modeling in the banking industry. The \textsc{loan} dataset (available on Kaggle) holds information on consumer loans issued by the Lending Club, a peer-to-peer lender. The raw dataset holds data on over 450,000 consumer loans issued between 2007 and 2014 with 73 features and one target variable. After pre-processing,
the final dataset consists of 395,492 samples with 42 features, including one-hot encoded features for categorical variables. The target variable is binary and indicates good or bad credit risk. Since the distribution of the target variable is highly imbalanced, with only 10\% of instances belonging to the positive class, we focus on both accuracy and F1-score as performance indicators. We look at the average rule length (``Avg. RL''), the number of rules (``NoR''), the average rule length per sample (``Avg. RLpS''), and the average number of rules per sample (``Avg. NoRpS'') as before. 

For consistency, we first focus on interpretability only and compare the results of RUG with results obtained with FSDT \citep{McTavish-etal-2021}, CG \citep{Lawless-etal-2021FairBoolean} and BinOCT \citep{VerwerZhang2019Learning}, and DT-h \citep{PATEL2024106579}.
Since shorter rules are considered to be more interpretable, we set the parameter for the maximum tree depth to two for FSDT, BinOCT, DT-h and RUG. For RUG, we set the cost coefficients $c_j$'s to be equal to the rule length, \textit{i.e.}, number of conditions in the rule. Our results are reported in Table \ref{tab:loan_interpretability_a}. RUG generates only seven rules, with which we can reach an accuracy of 97.55\% and an F1-score of 87.50\%. On average, 1.25 rules are used per sample to predict the target. With \texttt{max\_depth} set to two, the average rule length as well as the average rule length per sample is also equal to two. Both FSDT and DT-h, with four rules, achieve a relatively good accuracy of 95.72\% and 95.65\% with an F1-score of only 81.41\% and 81.37\%, respectively. BinOCT performs quite poorly with a F1-score of 0.00\% and an accuracy of 89.95\%. The complexity hyperparameter $C$ of CG needed to be set to five for it to return at least one rule. CG shows worse performance than both FSDT and RUG with an accuracy of 92.24\% and a much lower F1-score of only 39.47\%. 

\captionsetup[table]{labelfont=small, textfont=small}
\captionsetup[subtable]{labelfont=small, textfont=small}
{\renewcommand{\arraystretch}{0.8}
\begin{table}[ht]
\caption{Interpretability case study results}
\begin{subtable}[h]{\textwidth}
\centering
\caption{Results on the test set obtained with RUG, FSDT, BinOCT, CG, and DT-h when the focus is on interpretability.}
\label{tab:loan_interpretability_a}
\resizebox{0.7\textwidth}{!}{%
\begin{tabular}{@{}lrrrrrr@{}}
\hline \hline 
& {\textbf{F1 (\%)}} & {\textbf{Acc. (\%)}} & {\textbf{NoR}} & {\textbf{Avg.RL}} & {\textbf{Avg.NoRpS}} & \textbf{Avg.RLpS}\\ \hline
 RUG    &  87.50      & 97.55                               &   7.00        & 2.00 & 1.25 & 2.00             \\
FSDT & 81.41 & 95.72 & 4.00      & 2.00       & 1.00     & 2.00 \\
BinOCT & 0.00 & 89.95 & 4.00 & 2.00 & 1.00 & 2.00 \\
CG & 39.47 & 92.24 & 1.00 & 4.00 & 1.00 & 4.00 \\
  DT-h &   81.37 &   95.65 &   4.00 &   2.00 &   1.00 &   2.00\\
\hline \hline 
\multicolumn{7}{l}{\footnotesize {Avg.RL:} Average rule length; {NoR:} Number of rules; {Avg.NoRpS:} Average number of rules per}  \\ [-2mm]
\multicolumn{7}{l}{\footnotesize {sample; Avg.RLpS:} Average rule length per sample} \\ [-2mm]
\\
\end{tabular}%
}
\end{subtable}

\begin{subtable}[h]{\textwidth}
\centering
\caption{Results on the test set obtained with RUG, FSDT, BinOCT, CG, and DT-h when the focus is on accuracy.}
\label{tab:loan_interpretability_b}
\resizebox{0.7\textwidth}{!}{%
\begin{tabular}{@{}lrrrrrr@{}}
\hline \hline 
& {\textbf{F1 (\%)}} & {\textbf{Acc. (\%)}} & {\textbf{NoR}} & {\textbf{Avg.RL}} & {\textbf{Avg.NoRpS}} & \textbf{Avg.RLpS}\\ \hline
 RUG    &  94.68    & 98.96                                  &   44.00        & 1.98 & 6.55 & 2.00             \\
FSDT & 85.56 & 97.23 & 8.00 & 3.00 & 1.00 & 3.00 \\
BinOCT & -- & -- & -- & -- & -- & -- \\
CG & 59.05 & 93.89 & 4.00 & 4.50 & 1.35 & 4.95 \\
   DT-h &    92.94 &    98.58 &    256.00 &    8.00 &    1.00 &    8.00 \\
\hline \hline 
\multicolumn{7}{l}{\footnotesize {Avg.RL:} Average rule length; {NoR:} Number of rules; {Avg.NoRpS:} Average number of rules per}  \\ [-2mm]
\multicolumn{7}{l}{\footnotesize {sample; Avg.RLpS:} Average rule length per sample} \\ [-2mm]
\end{tabular}%
}
\end{subtable}
\end{table}}

Next, we try to boost the performance of all methods by relaxing some parameters, while still keeping the models interpretable. To that end, we allow higher \texttt{max\_depth} values for the trees of the FSDT, and the complexity parameter $C$ of the CG as well as the number of iterations in RUG. For the FSDT, the best value for the depth of the trees was three; for the CG the best value for the complexity parameter was 25. The accuracy and F1-score of DT-h were best for a depth of eight, although an even higher performance may be reached with a higher value for the maximum depth parameter, which however lead to execution failure potentially due to hardware limitations. BinOCT struggled with a higher depth parameter and a dataset of this size, and execution failed. This may of course also be due to limitations imposed by the hardware used to run these experiments. We report these results in Table \ref{tab:loan_interpretability_b}. RUG returns 44 rules with an average length of 1.98. The accuracy reaches 98.96\% and the F1-score increases to 94.68\% while maintaining a good level of local interpretability with an average of 6.55 rules per sample. Both CG and FSDT have better global interpretability considering their fewer rules, however, their accuracy does not match that of RUG and the F1-scores are lower with 85.56\% for FSDT and 59.05\% for CG. The performance of DT-h almost matches that of RUG, with an accuracy of 98.58\% and a F1-score of 92.94\%, however with a much more complex model consisting 256 rules each of length eight.

\begin{figure}[!ht]
    \centering
    \includegraphics[width=0.95\textwidth]{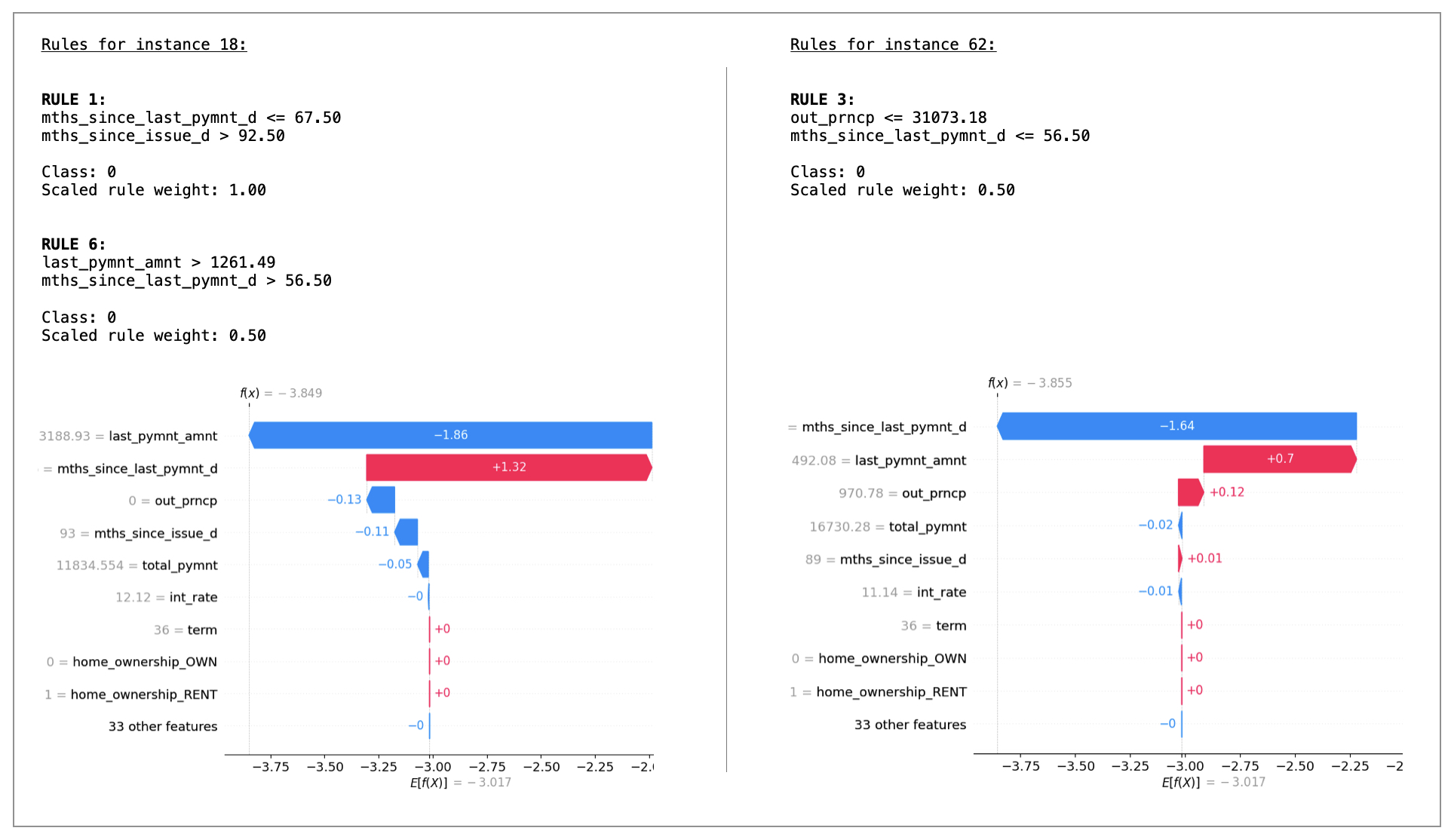}
    \caption{Local interpretations for two samples (left and right). The rules used by RUG to classify the samples are given above the feature importance plots that are produced after applying SHAP to the trained LightGBM model.}
    \label{fig:shap}
\end{figure}

To evaluate the performance of RUG further, we also apply SHAP \citep{LundbergLee2017SHAP} to LightGBM that we train on the dataset and compare its output to that of RUG (from Table \ref{tab:loan_interpretability_a}). With 80 rules, an average rule length of 2, and 20 rules on average per sample, the trained LightGBM model has a F1-score of 87.06\% and an accuracy of 97.27\%.
SHAP produces local explanations per sample in terms of feature weights and associates feature importance plots. Typically, in a feature importance plot, the features are sorted in descending order of the magnitudes of their weights, and the signs of the weights are depicted with two different colors. We have compared the SHAP output against the rules and the associated features obtained by RUG. In Figure \ref{fig:shap}, we display the SHAP plots for two rejected samples (class 0) alongside the rules generated by RUG that cover those samples. The first sample (Figure \ref{fig:shap} left) is covered by two rules, rule 1 and rule 6, which use features called as \texttt{last\_pymnt\_amnt}, \texttt{mths\_since\_issue\_d}, and \texttt{mths\_since\_last\_pymnt\_d}. These features are also among the four most important according to SHAP. The second sample (Figure \ref{fig:shap} right) is covered by just one rule, rule 3. Similarly, the features utilized by this rule belong to the features with the highest three contributing to the prediction of the LightGBM model. This exemplifies the advantage of using RUG for local explanations as an in-processing method. Indeed, RUG does not need an additional explanation step such as SHAP, being a post-processing method which requires substantial computational effort to yield an output.

In the Supplementary Material, we present further results on the case study where we compare RUG to traditional machine learning models including Logistic Regression (LR), DT, RF, ADA, and LightGBM. Besides, we introduce an implementation of RUG in practice at a consultancy firm.

\section{Conclusion}
\label{section::conclusion}

We have proposed a Linear Programming (LP) based rule generation framework to discover a set of rules for multi-class classification problems. The proposed LP-based framework is more advantageous than existing mathematical optimization-based approaches (MOBAs) that use mixed-integer linear programming (MILP) due to its scalability for large datasets. Our methodology is flexible to address interpretability and fairness both of which are salient to support the trustworthiness of algorithms in automated decision making.

The objective of our LP formulation is to find a set of rules and their associated weights such that the sum of the classification error and the total rule cost is minimized. The rule costs can promote shorter rules in the LP model for better interpretability. Optimal rule weights from the LP model facilitate interpretation for classification by attaching importance to the rules. 

Fairness often arises in the form of constraints to eliminate discrimination among protected and unprotected social groups in society. Most of the fairness definitions are designed for binary classification problems. We introduce two fairness notions that address multiple classes and a sensitive attribute (with multiple protected groups) for group fairness. 

We solve our LP using column generation. A significant advantage of this is having more control over the generation of the rules. We argue that it is still challenging to optimally solve the pricing subproblem as we have shown it to be NP-hard. As a remedy, we have proposed to solve a proxy subproblem by training DTs with sample weights. These sample weights are obtained from the dual of the linear programs. In fact, training DTs with sample weights are very fast, and also standard in most machine-learning packages. Such a column generation approach combined with solving the pricing subproblem quickly brings further scalability. 

We have demonstrated the performance of the proposed methodology using standard test instances from the literature as well as a case study. The numerical experiments have shown that our framework can be used to compromise the balance among accuracy, interpretability, and fairness. The case study is concerned with credit risk scoring in the banking industry and addresses interpretability.

We observe that our rule generation algorithm, RUG is highly scalable for large datasets unlike other MOBAs considered. RUG outperforms other methods in terms of both accuracy and F1-scores for the majority of the instances. Benchmark MOBAs perform very well for global interpretability. Nevertheless, RUG focuses on local interpretability with an aim to provide individual explanations for which those benchmark studies struggle since every individual is mostly treated using the same few rules. 

We confined the focus of our work on MOBAs, however it may be valuable to inspect and compare performance of RUG against off-the-shelf methods such as RF or AdaBoost. While these methods are inherently black-box (not interpretable) due to their number of estimators, a comparison in terms of fairness could be interesting, especially as there are approaches that incorporate fairness into existing machine learning methods \citep[e.g.,][]{HUANG2022117240,PessachShmueli2022FairnessReview}. 

This study is conducted for multi-class classification. As further research, our linear programming model can be reformulated for regression problems using a piecewise-linear loss function (\textit{e.g.}, mean absolute deviation). This requires analyzing the resulting pricing subproblems and the usage of new heuristic approaches if possible.  
Last but not least, robustness of the generated rules can be considered as a fruitful future research direction.

\section*{Acknowledgements}
The authors are thankful to Emre Can Yayla for his help with the preparation of datasets.

\clearpage
\appendix

\section{Proof of Proposition \ref{lem:psp}} \label{app::hardness_proof}
 
We show that the PSP represented with \eqref{PSP_explicit2} is NP-Hard by reduction from Maximal Coverage Location Problem (MCLP) presented in \cite{ChurchReelleMCP1974}. The proof consists of three parts: \textit{i}) reformulation of the PSP represented with \eqref{PSP_explicit2} as an equivalent mathematical program that can be done in polynomial time, \textit{ii}) introduction of the MCLP formulation that reduces to an instance of the PSP, and \textit{iii}) transformation of their solutions to each other. In what follows, we present these three parts in detail.
 
\textit{i}) 
We derive an equivalent formulation for the PSP. To that end, we replace the second constraint set in \eqref{PSP_explicit2} with
\begin{equation}
  \label{PSP_weak}
  \begin{array}{lll}
    & \frac{1}{|\CP_i|} \sum_{p \in \CP_i} z_p \leq b_i & \hspace{2.5cm} i \in \CI,
  \end{array}
\end{equation}
which can be obtained by summing over the second constraints of \eqref{PSP_explicit2} for $p \in \CP_i$. 
Lastly, the PSP given by \eqref{PSP_explicit2} is equivalently represented as follows:
\begin{equation}
  \label{PSP_explicit3}
  \begin{array}{lll}
  \maximize & \sum_{i \in \CI^-} b_i \beta_i -\sum_{i \in \CI^+} b_i \beta_i + C \\[2mm]
    \subto & \sum_{p \in \CP_i} z_p \geq b_i & i \in \CI, \\[2mm]
    & \frac{1}{|\CP_i|} \sum_{p \in \CP_i} z_p \leq b_i & i \in \CI, \\[2mm]
    & \sum_{p \in \CP} z_p = \ell, & \\[2mm]
    & b_i \in \{ 0, 1 \}, & i \in \CI, \\[2mm]
    & z_p \in \{ 0, 1 \}, & p \in \CP.
  \end{array}
\end{equation}

\textit{ii}) Now, we introduce the MCLP formulation \citep{ChurchReelleMCP1974} that is used for reduction:
\begin{equation}
  \label{PSP_MCLP}
  \max \left \{ \sum_{h \in \mathcal{H} }A_h Y_h :  \sum_{l \in \CL_h} X_l \geq Y_h, h \in \mathcal{H} ; \sum_{l \in \CL} X_l = T; X \in \{0,1\}^{|\mathcal{L}|}, Y \in \{0,1\}^{|\mathcal{H}|} \right \}.
\end{equation}
The objective of the MCLP is to maximize the weighted number of $h \in \mathcal{H}$ demand nodes having weight of $A_h \geq 0$ where $Y_h$ denotes the binary variable showing whether demand node $h$ is covered or not. Here, $\CL$ is the set of facility nodes and $X_l$ stands for the binary variable to open facility $l \in \CL$; $T$ is the number of facility nodes to open given in the second constraint as $\sum_{l \in \CL} X_l = T$; $\CL_h$ is the set of facility nodes that can serve demand node $h$, and thus, the first set of constraints is the coverage constraints for each demand node $h$. 

We solve a PSP instance with $\beta_i = 0$ for $i \in \CI^+$, and for each $i \in \CI$, define a demand node $h$ with $A_h = \beta_i$. We further define for each feature $p \in \CP$, a facility node $l$, and let $\CP_i$ be the set of demand nodes $h$ that can be served by facility node $l \in \CL_h$. Also, we set $\ell = T$ as the number of facility nodes to open. 

\textit{iii}) Lastly, we show how to transform solutions of the PSP and the MCLP to each other. Given an optimal solution $(\mathbf{b}^*, \mathbf{z}^*)$ of the PSP instance, $(\mathbf{b}^*, \mathbf{z}^*)$ is also optimal for the MCLP. Observe that, by selection of $\beta_i \geq 0$ values constraints $\frac{1}{|\CP_i|} \sum_{p \in \CP_i} z_p \leq b_i$ are redundant for $i \in \CI^{-}$ since a sample $i \in \CI^{-}$ is enforced to one as long as $\sum_{p \in \CP_i} z_p > 0 $. Besides, samples $i \in \CI^{+}$ has no effect on the objective as $\beta_i = 0$ which renders both of the following constraints $\sum_{p \in \CP_i} z_p \geq b_i$ and $\frac{1}{|\CP_i|} \sum_{p \in \CP_i} z_p \leq b_i$ redundant for $ i \in \CI^{+}$. Thus, $(\mathbf{b}^*, \mathbf{z}^*)$ is also optimal for the MCLP. Assume that $(\mathbf{b}^*, \mathbf{z}^*)$ is not optimal for the MCLP, then there exists a solution $(\bar{\mathbf{b}}^*, \bar{\mathbf{z}}^*)$ with a better objective. We can set $b_i = \bar{b}^{*}_i$ for $i \in \CI^{-}$. For each sample $i \in \CI^{+}$, we can check in polynomial time to find binding constraints in \eqref{PSP_explicit3}. We can set $b_i = 0$ when $\sum_{p \in \CP_i} z_p = 0$, and otherwise, set $b_i = 1$ without changing the objective value of the PSP. This implies that we can find a better objective than $(\mathbf{b}^*, \mathbf{z}^*)$ leading to a contradiction. This implies that the PSP is NP-complete by reduction from the MCLP, which is NP-hard as shown by \cite{Downs1991}. Consequently, the PSP represented with \eqref{PSP_explicit2} is also NP-hard.

\section{Fairness Discussion}\label{app:fairness}

In this section, we discuss relevant fairness notions from the literature for both binary and multi-class classification. Our fairness definitions introduced in section \ref{section::sub::interpretability} build upon this prior work.

\paragraph{Binary Classification} Let $y \in \{+1, -1\}$ stand for the labels of the samples having a positive ($+1$) or negative label ($-1$), and $\hat{y} \in \{+1, -1\}$ is the prediction of a classifier. Then, the total numbers of samples are represented as True Positive (TP) when $y = \hat{y} = +1$, as False Positive (FP) when $y = -1, \hat{y} = +1$, as False Negative (FN) when $y = +1, \hat{y} = -1$, and as True Negative (TN) when $y = \hat{y} = -1$. These numbers are then used by performance measures of binary classifiers.

Let $\mathcal{G}$ be the set of protected groups $g \in \mathcal{G}$ having distinct sensitive characteristics. Consider $g \in \mathcal{G} \equiv \{ 0,1\}$ showing that there are two protected groups, for instance, based on gender, where $g = 1$ and $g =0$ represent female and male individuals, respectively. The fairness notions are established using empirical probabilities, which are then transformed into fairness constraints. \cite{Hardt-etal-2016EqualOppOdd} define \textit{Equalized Odds} (EOD) and its relaxation \textit{Equal Opportunity} (EOP) for binary classification as follows: 
\begin{align}
   & \text{\it Equalized Odds:} && P(\hat{y} = +1 | y = a, g = 1) = P(\hat{y} = +1 | y = a, g = 0), \hspace{0.3cm} a \in \{+1, -1\},\label{eq:eq_odds} \\ 
   & \text{\it Equal Opportunity:} && P(\hat{y} = +1 | y = +1 , g = 1) = P(\hat{y} = +1 | y = +1 , g = 0). \label{eq:eq_opp}
\end{align}
These equalities indicate that the probability of predicting a positive class should be equal for both protected groups. EOD considers both TP and FP, whereas EOP addresses only TP. In fact, \cite{Lawless-etal-2021FairBoolean} define EOP and EOD differently such that EOD takes into account FP and FN while EOP seeks for equality of FN among protected groups. When we swap the labels of samples with each other in \cite{Lawless-etal-2021FairBoolean}, the EOP concept becomes identical in both studies \citep{Hardt-etal-2016EqualOppOdd,Lawless-etal-2021FairBoolean}. Unfortunately, a similar conclusion does not hold for the fairness notion using EOD. Akin to EOD by \cite{Lawless-etal-2021FairBoolean}, \textit{error rate balance} (ERB) is used as the fairness notion by \cite{Chouldechova2017FairPrediction} to equalize both error rates FP and FN among protected groups. Recently, \cite{Berk-etal-2021Fairness} name ERB as \textit{conditional procedure error} and overall error as \textit{overall procedure error} (OPE). To sum up, there is an abundance of fairness notions even for the binary classification, and consequently, it is fair to say that fairness terminology is not settled yet. 
Here, we concentrate on \textit{separation} criterion which implies that given a class of observations, the prediction is independent of the sensitive attribute as defined in \cite{FairMLBook_2019}. This reduces to equality of error rates among protected groups.

\paragraph{Multi-Class Classification}Fairness definitions start to levitate in a multi-class setting. The metrics TP, FP, TN, and FN require a positive and a negative class. Then, the consideration of other classes follows and multi-class setting eviscerates the meaning of TP, FP, FN, and TN. \cite{Denis-etal-2021MClassFairness} generalize \textit{demographic parity} (DP) definition into multi-class such that DP imposes equal prediction probability of each class among the protected groups. DP is being criticized by \cite{Dwork-etal-2012Fairness} that it might yield unfair outcomes for individuals while trying to guarantee equality among protected groups. \cite{Alghamdi-etal-2022MultiClassFairness} modify DP, EOD, and \textit{overall accuracy equality} (OAE) into a multi-class multiple protected groups setting by post-processing whereas our methodology belongs to in-processing techniques \citep[see][]{Wan-etal-2023In-Processing}. Observe that, maximizing overall accuracy corresponds to minimizing overall error. In that sense, they can be considered as complementing each other. \cite{Zafar-etal-2019FairnessFlex} use the term \textit{disparate mistreatment} (DM) where misclassification rate among protected groups are equalized for binary classification. We also denominate our fairness notions using a similar terminology with \cite{Zafar-etal-2019FairnessFlex} and choose DM in our definitions.

\newpage 
\section{Numerical Experiments}\label{app:exp}

In this section, we provide more details on the datasets used for our numerical experiments. We would like to point the reader to \cite{fabris2022algorithmic} for an overview and documentation of popular fairness datasets, especially \textsc{adult} and \textsc{compas}. Notably, we chose to exclude the well-known German Credit dataset due to substantial criticisms regarding the coding of sensitive attributes \citep{fabris2022algorithmic}.

\subsection{Dataset properties}\label{app:dataset_properties}

{\renewcommand{\arraystretch}{0.55}
\captionsetup[table]{labelfont=small, textfont=small}
\begin{table}[!h]
  \caption{Properties of the datasets used for the numerical experiments on interpretability.}
  \label{table:dataset} 
  \vspace{-0.2cm}
\begin{center}
\begin{small}
    \begin{tabular*}{\textwidth}{@{\extracolsep{\fill}}lrrrrrr}
 
      \hline\hline
       & \multirow{2}{*}{\textbf{Sample Size}}	 & 	\multirow{2}{*}{\textbf{Classes}} & \multirow{2}{*}{\textbf{Features}}	  & \textbf{Encoded} & \textbf{Binarized} & \textbf{Imbalance}\\ 
       & 	 & 	&  & \textbf{features} & \textbf{features} & \textbf{(\%)} \\
      \hline
\textsc{adult} & 	32561 & 	2 & 	14 & 107 & 262 & 24.08 \\
\textsc{bank-mkt} & 	11162 & 	2 & 	16 & 51 & 180 & 47.38 \\
\textsc{banknote} & 	1372 & 	2 & 	4  & 4 & 72 & 44.46 \\
\textsc{diabetes} & 	768 & 	2 & 	8 & 8 & 134 & 34.90 \\
\textsc{hearts} & 	303 & 	2 & 	13 & 30 & 134 & 54.46 \\
\textsc{ilpd} & 	583 & 	2 & 	10 & 10 & 158 & 28.50  \\
\textsc{ionosphere} & 	351 & 	2 & 	34 & 34 & 566 & 64.10 \\
\textsc{liver} & 	345 & 	2 & 	6 & 6 & 104 & 57.97 \\
\textsc{loan} & 395492 & 2 & 21 & 42 & 302 & 10.05 \\
\textsc{magic} & 	19020 & 	2 & 	10 & 10 & 180 & 64.84 \\
\textsc{mammography} & 	11183 & 	2 & 	6 & 6 & 70 & 2.32 \\
\textsc{mushroom} & 	8124 & 	2 & 	22 & 117 & 234 & 51.80 \\
\textsc{musk} & 	6598 & 	2 & 	166 & 166 & 2922 & 15.41 \\
\textsc{oilspill} & 	937 & 	2 & 	49 & 48 & 772 & 4.38 \\
\textsc{phoneme} & 	5404 & 	2 & 	5 & 5 & 90 & 29.35 \\
\textsc{skinnonskin} & 	245057 & 	2 & 	3 & 3 & 54 & 79.25 \\
\textsc{tic-tac-toe} & 	958 & 	2 & 	9 & 27 & 54 & 65.34\\
\textsc{transfusion} & 	748 & 	2 & 	4 & 4 & 64 & 23.80 \\
\textsc{wdbc} & 	569 & 	2 & 	30 & 30 & 540 & 37.26 \\
\textsc{ecoli} & 	336 & 	8 & 	7 & 7 & -- &  -- \\
\textsc{glass} & 	214 & 	6 & 	9  & 9 & -- & --\\
\textsc{seeds} & 	210 & 	3 & 	7  & 7 & -- & --\\
\textsc{sensorless} & 	58509 & 	11 & 	48 & 48 & -- &  -- \\
\textsc{wine} & 	178 & 	3 & 	13  & 13 & -- & --\\
      \hline\hline
\multicolumn{7}{l}{\scriptsize \textsc{phoneme} is from an online repository: https://datahub.io/machine-learning/phoneme; \textsc{oilspill} comes from} \\
\multicolumn{7}{l}{\scriptsize \cite{Kubat1998_Oilspill_dataset}; All remaining datasets come from \cite{UCI-Repository2019}} \\
    \end{tabular*}
\end{small}
\end{center}
\end{table}

\begin{table}[!ht]
  \caption{\small{Properties of the datasets used for the numerical experiments for fairness.}} \label{table:datasets-fairness} 
  \vspace{-0.2cm}
\begin{center}
\begin{small}
  \resizebox{\textwidth}{!}{%
    \begin{tabular}{@{}lrrrrrrr@{}}
      \hline\hline
       & 	\multirow{2}{*}{\textbf{Sample Size}} & 	\multirow{2}{*}{\textbf{Classes}} & 	\multirow{2}{*}{\textbf{Features}} & \textbf{Encoded} & \textbf{Binarized} & \textbf{Sensitive} & \textbf{Groups}\\ 
       & & & &  \textbf{features} & \textbf{features} & \textbf{attribute}\\
      \hline
\textsc{adult} & 	32561 & 	2 & 	14 & 107 & 262 & Gender & Male/Female (2) \\
\textsc{compas} & 	6172 & 	2 & 	7 & 7 & 26 & Race & Caucasian/Other (2) \\
\textsc{default} & 	30000 & 	2 & 	23 & 32 & 316 & Gender & Male/Female (2) \\
\textsc{attrition} & 	1469 & 	2 & 	34 & 54 & -- & Work-life balance &  Bad/Good/Better/Best (4) \\
\textsc{law} & 	22386 &  5 & 5 & 5 & -- & Race	& White/Other (2) \\
\textsc{nursery} & 	12960 & 	5 & 	8 & 24 & -- & Parents' occupation & Usual/Pretentious/Great\_pret (3) \\
\textsc{student} & 	649 & 	5 & 	32 & 113 & -- & Address & Rural/Urban (2) \\

      \hline\hline
\multicolumn{8}{l}{\footnotesize \textsc{attrition}, \textsc{law}, \textsc{nursery}, and \textsc{student} do not require a binary version as they are only run with (Fair)RUG.}
\end{tabular}
}
\end{small}
\end{center}
\end{table}}
\newpage
\subsection{Further Results on the Comparison of RUG with Proxy PSP and RUG with Exact PSP}\label{app:proxy_exact}

\captionsetup[longtable]{labelfont=small, textfont=small, font=rm}
{\renewcommand{\arraystretch}{0.6}
\renewcommand\familydefault\sfdefault
\small
\begin{longtable}[!ht]{@{}llrrr@{}}
\caption{Results of RUG (h) vs. the exact formulation (e).}
\label{tab:results-rug-exact-short}\\
\hline \hline
                            &   & \textbf{CPU (s)}        & \textbf{F1 (\%)} & \textbf{Acc. (\%)} \\* \midrule
\endfirsthead
\multicolumn{4}{c}%
{ Table \thetable\ continued from previous page} \\
\toprule
                            &             & \textbf{F1 (\%)} & \textbf{Acc. (\%)} \\* \midrule
\endhead
\bottomrule
\endfoot
\endlastfoot
%
\multirow{2}{*}{\textsc{adult}} & h & 7.67 & 62.30 & 83.56 \\
& e &300.00 & 58.75 & 82.67 \\*
\midrule
\multirow{2}{*}{\textsc{bank-mkt}} & h &3.52 & 85.28 & 85.67 \\
& e &300.00  & 80.15 & 80.79 \\*
\midrule
\multirow{2}{*}{\textsc{banknote}} & h & 0.27 & 100.00 & 100.00 \\
& e &300.00  & 99.19 & 99.27 \\*
\midrule
\multirow{2}{*}{\textsc{diabetes}} & h & 0.16 & 44.71 & 69.48 \\
& e &300.00  & 45.00 & 71.43 \\*
\midrule
\multirow{2}{*}{\textsc{hearts}} & h & 0.01  & 100.00 & 100.00 \\
& e & 207.85  & 100.00 & 100.00 \\*
\midrule
\multirow{2}{*}{\textsc{ilpd}} & h & 0.21  & 36.62 & 61.21 \\
& e &300.00  & 36.36 & 63.79 \\*
\midrule
\multirow{2}{*}{\textsc{ionosphere}} & h & 0.08  & 92.13 & 90.14 \\
& e &300.00  & 92.47 & 90.14 \\*
\midrule
\multirow{2}{*}{\textsc{liver}} & h & 0.13 & 56.41 & 50.72 \\
& e &300.00  & 62.34 & 57.97 \\*
\hline
\multirow{2}{*}{\textsc{magic}} & h & 6.01  & 89.83 & 86.33 \\
& e &300.00  & 87.99 & 83.39 \\*
\hline
\multirow{2}{*}{\textsc{mammography}} & h & 0.77 & 53.16 & 98.35 \\ 
 & e &300.00  & 54.32 & 98.35\\*
\hline
\multirow{2}{*}{\textsc{mushroom}} & h & 0.35  & 100.00 & 100.00 \\
 & e &300.00  & 99.94 & 99.94 \\*
\hline
\multirow{2}{*}{\textsc{musk}} & h & 3.52  & 84.70 & 95.76 \\
& e &300.00  & 75.64 & 93.56 \\*
\hline
\multirow{2}{*}{\textsc{oilspill}} & h & 0.15 & 37.50 & 94.68 \\
& e &300.00  & 31.58 & 93.09 \\*
\hline
\multirow{2}{*}{\textsc{phoneme}} & h & 1.58 & 79.75 & 88.07 \\
& e &300.00  & 65.38 & 81.78 \\*
\hline
\multirow{2}{*}{\textsc{skinnonskin}} & h & 31.16  & 99.40 & 99.05 \\
 & e &300.00  & 98.69 & 97.93 \\*
\hline
\multirow{2}{*}{\textsc{tictactoe}} & h & 0.28  & 98.43 & 97.92 \\
& e &300.00  & 95.97 & 94.79 \\*
\midrule
\multirow{2}{*}{\textsc{transfusion}} & h & 0.06  & 46.15 & 76.67 \\
 & e &300.00  & 00.00 & 76.00 \\*
 \midrule
\multirow{2}{*}{\textsc{wdbc}} & h & 0.14  & 95.12 & 96.49 \\
& e &300.00  & 93.67 & 95.61 \\*
\hline
\multirow{2}{*}{Average scores} & h & 3.12 & 75.64 & 87.40 \\
 & e & 294.88   & 70.97 & 86.69 \\*
\hline\hline
\multicolumn{4}{l}{\scriptsize Hyperparameter sets for heuristic: max\_depth = \{3,5\}, } \\
\multicolumn{4}{l}{\scriptsize pen\_par = \{0.1,1.0,10.0\} and max\_RMP\_calls = \{5,15,30\}}
\end{longtable}
}

\newpage
\subsection{Further Results on Interpretability}\label{app:interpretability}

\vspace{0.5cm}
\setlength{\LTcapwidth}{0.9\textwidth}
\captionsetup[longtable]{labelfont=small, textfont=small, font=rm}
{\renewcommand{\arraystretch}{0.55}
\renewcommand\familydefault\sfdefault
\small
\begin{longtable}{@{}p{0.17\textwidth}p{0.07\textwidth}R{0.11\textwidth}R{0.08\textwidth}R{0.10\textwidth}R{0.08\textwidth}R{0.06\textwidth}R{0.07\textwidth}R{0.07\textwidth}@{}}

\caption{Results of RUG, FSDT, BinOCT, and CG using five-fold cross validation and grid search for hyperparameter tuning on a hold-out test.}
\label{tab:results}\\
\hline \hline
                            &            & \textbf{CPU (s)} & \textbf{F1 (\%)} & \textbf{Acc. (\%)}  & \textbf{NoR} & \textbf{Avg. RL} & \textbf{Avg. NoRpS} & \textbf{Avg. RLpS} \\* \midrule
\endfirsthead
\multicolumn{9}{c}%
{Table \thetable\ continued from previous page} \\
\toprule
                            &            & \textbf{CPU (s)} & \textbf{F1 (\%)} & \textbf{Acc. (\%)} & \textbf{NoR} & \textbf{Avg. RL} & \textbf{Avg. NoRpS} & \textbf{Avg. RLpS} \\* \midrule
\endhead
\bottomrule
\endfoot
\endlastfoot
%
\multirow{5}{*}{\textsc{ADULT}} & RUG & 9.95 & 64.09 & 85.58 & 18.00 & 2.44 & 1.13 & 2.08 \\
& FSDT & 300.00 & 59.13 & 83.06 & 17.00 & 4.76 & 1.00 & 2.86 \\
& BinOCT*  & 300.00 & 0.00 & 75.93 & 8.00 & 3.00 & 1.00 & 3.00  \\
& CG & 300.00 & 46.30 & 80.41 & 4.00 & 5.00 & 1.13 & 5.00 \\
& DT-h & 176.73 & 67.62 & 85.11 & 1024.00 & 10.00 & 1.00 & 10.00 \\*
\midrule
\multirow{5}{*}{\textsc{BANK\_MKT}} & RUG & 3.90 & 86.58 & 86.83 & 110.00 & 3.97 & 9.06 & 4.03 \\
& FSDT & 83.23 & 77.36 & 78.64 & 8.00 & 3.00 & 1.00 & 3.00 \\
& BinOCT*     &  300.00 & 0.00 & 52.62 & 8.00 & 3.00 & 1.00 & 3.00 \\
& CG & 300.00 & 77.34 & 77.88 & 3.00 & 4.00 & 1.28 & 4.00 \\
& DT-h & 26.27 & 81.97 & 82.76 & 1024.00 & 10.00 & 1.00 & 10.00 \\* 
\midrule
\multirow{5}{*}{\textsc{BANKNOTE}} & RUG & 0.23 & 100.00 & 100.00 & 30.00 & 2.33 & 6.62 & 2.38 \\
& FSDT & 3.44 & 99.17 & 99.27 & 25.00 & 4.76 & 1.00 & 4.41 \\
& BinOCT     & 300.00 & 97.52 & 97.82 & 8.00 & 3.00 & 1.00 & 3.00 \\
& CG & 300.00 & 96.77 & 97.09 & 3.00 & 3.00 & 1.31 & 2.46 \\
& DT-h* & 300.00 & 98.78 & 98.91 & 1024.00 & 10.00 & 1.00 & 10.00 \\*
\midrule
\multirow{5}{*}{\textsc{DIABETES}} & RUG & 0.13 & 41.46 & 68.83 & 21.00 & 2.62 & 3.45 & 2.79 \\
& FSDT & 9.99 & 53.33 & 72.73 & 8.00 & 3.00 & 1.00 & 3.00 \\
& BinOCT  & 300.00 & 37.50 & 35.06 & 8.00 & 3.00 & 1.00 & 3.00 \\
& CG & 300.00 & 50.57 & 72.08 & 2.00 & 4.00 & 1.09 & 3.15 \\
& DT-h & 299.41 & 59.34 & 75.97 & 32.00 & 5.00 & 1.00 & 5.00 \\*
\midrule
\multirow{5}{*}{\textsc{HEARTS}} & RUG & 0.03 & 100.00 & 100.00 & 5.00 & 1.60 & 2.18 & 1.39 \\
& FSDT & 0.01 & 100.00 & 100.00 & 4.00 & 2.25 & 1.00 & 1.82 \\
& BinOCT  & 29.47 & 30.19 & 39.34 & 8.00 & 3.00 & 1.00 & 3.00 \\
& CG & 0.81 & 100.00 & 100.00 & 2.00 & 4.00 & 1.75 & 4.00 \\
& DT-h & 0.33 & 100.00 & 100.00 & 8.00 & 3.00 & 1.00 & 3.00 \\*
\midrule
\multirow{5}{*}{\textsc{ILPD}} & RUG & 0.13 & 22.64 & 64.66 & 23.00 & 3.13 & 2.19 & 3.31 \\
& FSDT & 300.00 & 39.29 & 70.69 & 30.00 & 8.73 & 1.00 & 5.02 \\
& BinOCT     & 300.00 & 49.52 & 54.31 & 8.00 & 3.00 & 1.00 & 3.00 \\
& CG & 300.00 & 35.71 & 68.97 & 2.00 & 4.00 & 1.00 & 4.00 \\
& DT-h & 299.97 & 38.10 & 66.38 & 8.00 & 3.00 & 1.00 & 3.00 \\
\midrule
\multirow{5}{*}{\textsc{IONOSPHERE}} & RUG & 0.07 & 95.74 & 94.37 & 13.00 & 2.15 & 5.01 & 2.61 \\
 & FSDT & 300.00 & 90.53 & 87.32 & 13.00 & 4.38 & 1.00 & 4.30 \\
 & BinOCT     & 300.00 & 31.58 & 26.76 & 8.00 & 3.00 & 1.00 & 3.00 \\
& CG & 300.00 & 93.48 & 91.55 & 2.00 & 5.00 & 1.17 & 5.00 \\
& DT-h* & 64.49 & 90.91 & 88.73 & 1024 & 10.00 & 1.00 & 10.00 \\*
\hline
\multirow{5}{*}{\textsc{LIVER}} & RUG & 0.11 & 56.41 & 50.72 & 33.00 & 3.27 & 3.87 & 3.28 \\
 & FSDT & 300.00 & 70.59 & 63.77 & 31.00 & 4.97 & 1.00 & 4.97 \\ 
& BinOCT     & 300.00 & 58.82 & 59.42 & 8.00 & 3.00 & 1.00 & 3.00 \\
& CG & 300.00 & 68.89 & 59.42 & 2.00 & 2.50 & 1.02 & 2.21 \\ 
& DT-h & 62.00 & 61.54 & 56.52 & 1024.00 & 10.00 & 1.00 & 10.00 \\*
\hline
\multirow{5}{*}{\textsc{MAGIC}} & RUG & 5.48 & 90.17 & 86.78 & 125.00 & 3.90 & 5.61 & 3.99 \\
& FSDT & 132.74 & 86.97 & 82.02 & 8.00 & 3.00 & 1.00 & 3.00 \\
& BinOCT*     & 300.00 & 0.00 & 35.17 & 8.00 & 3.00 & 1.00 & 3.00 \\
& CG & 300.00 & 85.00 & 79.92 & 3.00 & 5.00 & 1.06 & 5.00 \\
& DT-h* & 264.84 & 88.48 & 84.04 & 32.00 & 5.00 & 1.00 & 5.00 \\*
\hline
\multirow{5}{*}{\textsc{MAMMOGRAPHY}} & RUG & 2.36 & 70.45 & 98.84 & 39.00 & 2.41 & 4.09 & 2.59 \\
 & FSDT & 4.09 & 47.22 & 98.30 & 8.00 & 3.00 & 1.00 & 3.00 \\
 & BinOCT     & 300.00 & 30.77 & 97.99 & 8.00 & 3.00 & 1.00 & 3.00 \\
& CG & 300.00 & 52.63 & 98.39 & 2.00 & 4.00 & 1.04 & 4.00 \\
& DT-h & 198.64 & 70.33 & 98.79 & 32.00 & 5.00 & 1.00 & 5.00 \\*
\hline
\multirow{5}{*}{\textsc{MUSHROOM}} & RUG & 0.46 & 100.00 & 100.00 & 10.00 & 2.70 & 3.70 & 2.86 \\
& FSDT & 0.02 &   100.00  & 100.00 & 13.00 & 4.08 & 1.00 & 3.94 \\
& BinOCT*     & 300.00 & 0.00 & 48.18 & 8.00 & 3.00 & 1.00 & 3.00 \\
& CG & 300.00 & 95.99 & 96.00 & 1.00 & 4.00 & 1.00 & 4.00 \\
& DT-h & 24.59 & 100.00 & 100.00 & 1024.00 & 10.00 & 1.00 & 10.00 \\*
\hline
\multirow{5}{*}{\textsc{MUSK}} & RUG & 1.93 & 87.77 & 96.52 & 61.00 & 4.18 & 5.10 & 4.43 \\
& FSDT & 300.00 & 64.07 & 90.91 & 8.00 & 3.00 & 1.00 & 3.00 \\
& BinOCT*     & 300.00 & 0.00 & 84.62 & 8.00 & 3.00 & 1.00 & 3.00 \\
& CG & 300.00 & 76.79 & 93.86 & 4.00 & 5.00 & 1.27 & 5.00 \\
& DT-h & 185.53 & 87.72 & 96.29 & 1024.00 & 10.00 & 1.00 & 10.00 \\*
\hline
\multirow{5}{*}{\textsc{OILSPILL}} & RUG & 0.21 & 62.50 & 96.81 & 32.00 & 2.66 & 5.90 & 2.96 \\
& FSDT & 300.00 & 50.00 & 96.81 & 8.00 & 3.00 & 1.00 & 3.00 \\
& BinOCT*     & 300.00 & 0.00 & 95.74 & 8.00 & 3.00 & 1.00 & 3.00 \\
& CG & 300.00 & 0.00 & 95.21 & 2.00 & 4.00 & 1.00 & 4.00 \\
& DT-h* & 296.93 & 60.00 & 95.74 & 32.00 & 5.00 & 1.00 & 5.00 \\*
\hline
\multirow{4}{*}{\textsc{PHONEME}} & RUG & 1.49 & 77.46 & 86.86 & 65.00 & 3.11 & 3.82 & 3.13 \\
& FSDT & 300.00 & 72.96 & 84.37 & 32.00 & 5.00 & 1.00 & 5.00 \\
& BinOCT*  & 300.00 & 66.98 & 80.57 & 8.00 & 3.00 & 1.00 & 3.00 \\
& CG & 300.00 & 63.41 & 81.31 & 3.00 & 4.33 & 1.20 & 4.24 \\
& DT-h & 232.34 & 74.88 & 85.66 & 1024.00 & 10.00 & 1.00 & 10.00 \\*
\hline
\multirow{5}{*}{\textsc{SKINNONSKIN}} & RUG & 41.13 & 99.94 & 99.91 & 103.00 & 2.57 & 6.95 & 2.31 \\
& FSDT & 300.00 & 99.14 & 98.64 & 25.00 & 4.76 & 1.00 & 4.50 \\
& BinOCT                       & --                    & --                  & --              & --              & --                & --                  & --                  \\
& CG & 300.00 & 95.53 & 92.83 & 3.00 & 2.67 & 1.33 & 2.32  \\
& DT-h* & 199.43 & 99.05 & 98.51 & 32.00 & 5.00 & 1.00 & 5.00 \\*
\hline
\multirow{5}{*}{\textsc{TICTACTOE}} & RUG & 0.32 & 98.04 & 97.40 & 35.00 & 4.20 & 2.61 & 4.09 \\
& FSDT & 300.00 & 89.33 & 85.94 & 27.00 &  4.85 & 1.00 & 4.51 \\
& BinOCT     & 300.00 & 58.13 & 55.73 & 8.00 & 3.00 & 1.00 & 3.00 \\
& CG & 300.00 & 90.43 & 88.54 & 9.00 & 3.44 & 1.63 & 3.28 \\
& DT-h & 173.80 & 94.21 & 92.19 & 1024.00 & 10.00 & 1.00 & 10.00 \\*
\hline
\multirow{5}{*}{\textsc{TRANSFUSION}} & RUG & 0.06 & 40.74 & 78.67 & 11.00 & 2.91 & 0.96 & 2.82 \\
 & FSDT & 0.71 & 48.48 & 77.33 & 8.00 & 3.00 & 1.00 & 3.00 \\
 & BinOCT     & 300.00 & 43.30 & 63.30 & 8.00 & 3.00 & 1.00 & 3.00 \\
& CG & 300.00 & 16.67 & 73.33 & 2.00 & 4.50 & 1.00 & 4.33 \\
& DT-h & 57.85 & 47.46 & 79.33 & 8.00 & 3.00 & 1.00 & 3.00 \\*
\hline
\multirow{5}{*}{\textsc{WDBC}} & RUG & 0.11 & 91.36 & 93.86 & 24.00 & 3.04 & 5.79 & 3.19 \\
& FSDT & 3.91 & 86.96 & 89.47 & 62.00 & 7.23 & 1.00 & 6.68 \\
& BinOCT*   & 300.00 & 95.24 & 96.49 & 8.00 & 3.00 & 1.00 & 3.00 \\
& CG & 300.00 & 91.36 & 93.86 & 3.00 & 4.00 & 1.54 & 4.59 \\*
& DT-h & 100.90 & 91.76 & 93.86 & 1024.00 & 10.00 & 1.00 & 10.00 \\*
\hline
 \multirow{4}{*}{\textsc{Average scores}}& RUG & 3.78 &	76.96 &	88.15 &	42.11 &	2.96 & 4.11 & 3.01 \\
& FSDT & 163.23 & 74.14 & 86.63 & 18.61 & 4.27 & 1.00 &	3.83\\
 & BinOCT     & 300.00 &	35.27	& 64.65	& 8.00 & 3.00 &	1.00 &	3.00 \\
& CG & 283.38 & 68.70 & 85.59 & 2.89 & 4.00 &	1.21 & 3.92 \\
& DT-h & 164.67	& 78.45 & 87.71 & 579.11 & 7.44 & 1.00 & 7.44 \\
\hline \hline
\multicolumn{9}{l}{\footnotesize Hyperparameter sets for RUG: max\_depth = \{3,5\}, pen\_par = \{0.1,1.0,10.0\}, max\_RMP\_calls = \{5,15,30\}} \\
\multicolumn{9}{l}{\footnotesize Hyperparameter sets for FSDT, BinOCT, CG, DT-h: max\_depth = \{3,5,10\}} \\
\multicolumn{9}{l}{\footnotesize The time limit was set to 300 seconds. If the time limit was hit we report the solution found within the limit.} \\
\multicolumn{9}{l}{\footnotesize -- indicates that the method did not return any solution within the time limit.}\\
\multicolumn{9}{l}{\footnotesize * hyperparameter sets excluded some parameters in cases where those lead to execution failure}
\end{longtable}
}

\clearpage
\bibliographystyle{elsarticle-harv} 
\bibliography{main}

\end{document}